\documentclass[runningheads]{llncs}

\usepackage{eccv}

\usepackage{eccvabbrv}

\usepackage{graphicx}
\usepackage{booktabs}
\usepackage{algorithm}
\usepackage{algpseudocode}
\usepackage[table]{xcolor}
\usepackage{multirow}
\definecolor{lightred}{RGB}{255,220,220} 
\definecolor{darkgreen}{RGB}{0, 100, 0}
\usepackage{wrapfig}

\usepackage[accsupp]{axessibility}  

\usepackage{hyperref}

\usepackage{orcidlink}
\usepackage{subcaption}

\begin{document}

\title{Semantically Aligned Gradient-Driven Context-Preserving Image Editing} 

\titlerunning{IABEdit}

\author{Chiranjeev Chiranjeev\inst{1}\orcidlink{0000-0003-3026-1255} \and
Muskan Dosi\inst{1}\orcidlink{0000-0001-7451-3317} \and
Mayank Vatsa\inst{1}\orcidlink{0000-0001-5952-2274} \and
Richa Singh\inst{1}\orcidlink{0000-0003-4060-4573}}

\authorrunning{Chiranjeev et al.}

\institute{Indian Institute of Technology Jodhpur, India
\email{\{chiranjeev.1,dosi.1,mvatsa,richa\}@iitj.ac.in} \\
\textcolor{magenta}{\url{https://github.com/IAB-IITJ/IABEdit}}}

\maketitle

\begin{center}
    \captionsetup{type=figure}
    \includegraphics[width=0.92\textwidth]{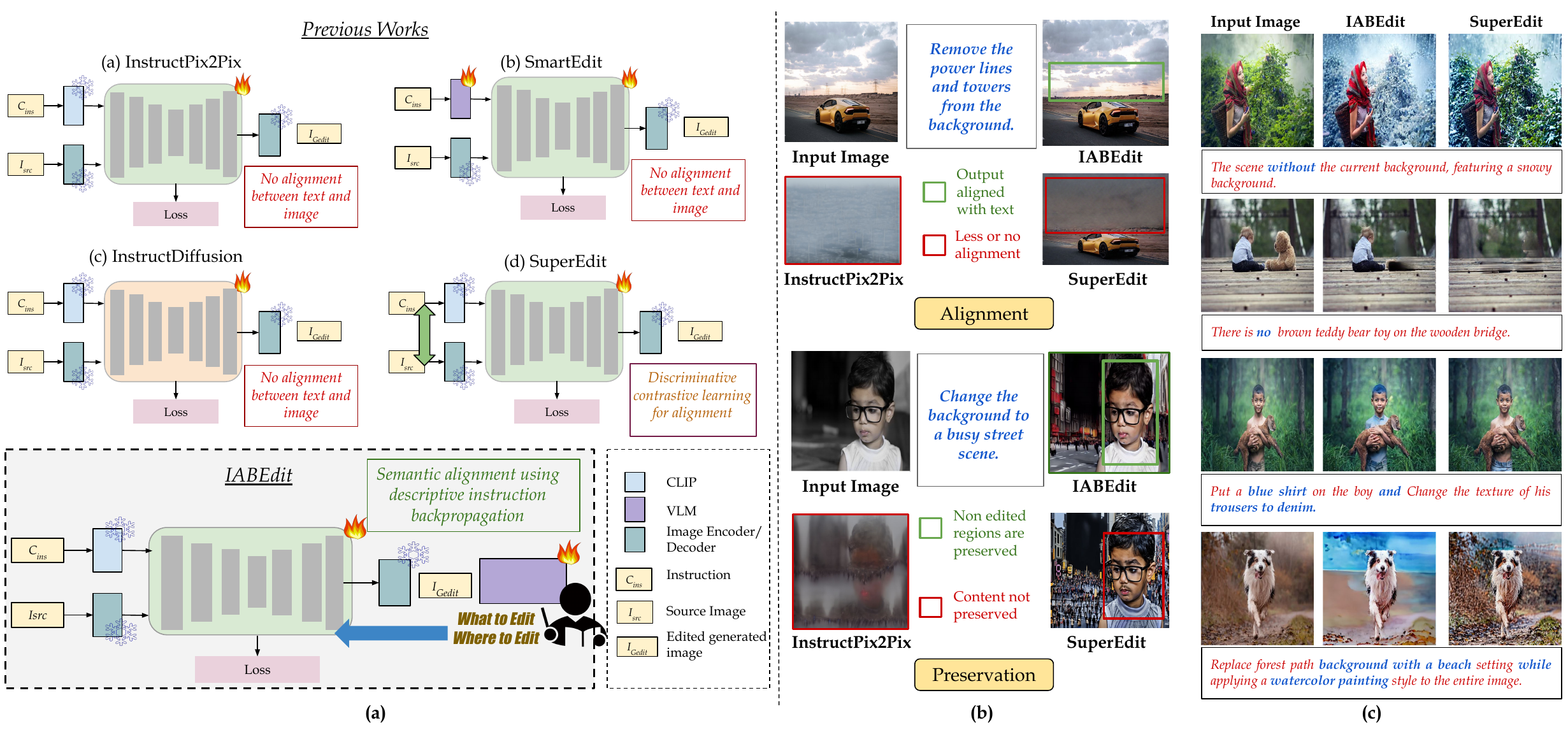}
    \captionof{figure}{(a) Existing editing models rely on static textual features, causing weak alignment and unwanted changes. (b) IABEdit improves both alignment and preservation through gradient-enabled semantic supervision. (c) Examples show IABEdit’s stronger semantic grounding and localized edits compared to prior methods.}
    \label{fig:vis_abstract}
\end{center}

\begin{abstract}

Instruction-guided image editing has a training-time blind spot. Generative editors are never required to semantically verify whether their outputs actually satisfy the instruction. Supervision stops at reconstruction and input textual-level conditioning. This produces incomplete edits, spatial spillover, and poor localization. We present \textit{IABEdit}, a model-agnostic framework that embeds differentiable semantic verification into training. A frozen vision-language model extracts spatially-aware descriptors from the ground-truth edit. A trainable aligner then reproduces them from the generated output. The residual between the two becomes a gradient that teaches the generator both what to edit and where, with no inference-time VLM cost. IABEdit is compatible with diverse backbones, including U-Net (Stable Diffusion) and MMDiT (FLUX), without altering their inference pipelines. On MagicBrush, it improves structural fidelity by +3.49 DINO-I over the best diffusion baseline and +1.26 over the best overall baseline, while remaining competitive on instruction alignment. It also achieves state-of-the-art instruction adherence performance on RealEdit and EMU Edit benchmarks based on embedding-based metrics. Most consequentially, on the D-LORD surveillance benchmark, it surpasses the proprietary Gemini agent by +5.13 DINO-P under heavy occlusion, where preserving identity is hardest. This shows that gradient-aligned VLM distillation holds up under real-world-like surveillance and occlusion conditions. Human and GPT-4o evaluations confirm perceptually precise, well-localized edits.


\end{abstract}

\section{Introduction}

Recent advances in generative image modeling, including diffusion models \cite{ho2020denoising, dhariwal2021diffusion, nichol2022glide, saharia2022photorealistic} with U-Net and transformer-based generators, and flow-matching frameworks such as FLUX \cite{labs2025flux} have enabled high-quality and flexible image manipulation. Instruction-guided image editing extends these capabilities by allowing users to modify images through natural language, supporting tasks such as object removal, attribute modification, scene transformation, and semantic style transfer. Earlier approaches relied on manually specified spatial masks, limiting scalability and usability. Modern generative models \cite{esser2024scaling, podell2024sdxl, ramesh2021zero, ramesh2022hierarchical, rombach2022high, labs2025flux} instead condition directly using textual instructions, enabling language-based editing.

Despite their realism, existing instruction-guided editors are weak in instruction–edit alignment as shown in Figure \ref{fig:vis_abstract}(b), and share a fundamental limitation: the training objective does not explicitly verify whether the generated image semantically satisfies the intended edit. While instructions influence the generation process, there is typically no mechanism that measures alignment between the desired transformation and the produced output beyond reconstruction or token-level supervision. Crucially, the model is never required to verify whether the progressively denoised image actually satisfies the instruction. As a result, this design leads to systematic failure modes: (1) incomplete semantic realization (under-editing), (2) spatial mislocalization of edits, and (3) unintended modification of non-target regions (over-editing).  The underlying issue is not architectural - it is the absence of explicit semantic supervision that enforces instructions.

We argue that instruction-guided editing should be formulated as a semantic alignment problem rather than purely a conditioning problem. To this end, we propose \textbf{\textit{IABEdit}}: \textit{\underline{I}nstruction}-\textit{\underline{A}ligned} \textit{\underline{B}ackpropagation} \textit{\underline{I}mage} \textit{\underline{Edit}ing}. IABEdit introduces differentiable semantic verification into the training of generative image editors. Instead of relying solely on textual conditioning for obtaining intended edits, we incorporate high-capacity vision-language reasoning to define what a successful edit should represent at a semantic level. The overview of IABEdit is shown in Figure \ref{fig:vis_abstract}(a).

This semantic supervisor evaluates whether the generated image reflects the intended transformation, capturing both global changes and localized attributes. By making this evaluation differentiable, the generative model is guided by an explicit measure of instruction satisfaction during optimization. Editing thus becomes a process of minimizing the semantic discrepancy between the intended outcome and the generated image. This training paradigm encourages the model to learn not only what must change, but also what must remain preserved, resulting in improved localization, stronger instruction adherence, and better structural consistency. IABEdit facilitates better alignment using supervision.

Importantly, IABEdit operates as an architecture-agnostic training-time supervision layer. It can be integrated into diffusion models, transformer-based generators, including MMDiTs in flow-matching frameworks, without altering their inference pipelines. Rich vision-language reasoning supervision is used only during training; at deployment (testing phase), editing remains a standard generative process with no additional computational overhead.
By reframing instruction-guided editing as semantic alignment optimization, IABEdit establishes a principled and scalable framework for semantically grounded image editing. Across diverse benchmarks and generative backbones, we demonstrate that enforcing semantic consistency during training yields more faithful, localized edits, independent of the underlying architecture. Our key contributions are:

\begin{enumerate}

    \item \textbf{Differentiable Semantic Reasoning-based Verification:} We introduce a gradient-based training objective that backpropagates VLM-derived semantic descriptors through the generative backbone, explicitly teaching both \textit{what} to edit and \textit{where} to edit. To the best of our knowledge, IABEdit is among the first methods to embed \textcolor{black}{textual semantic reasoning-oriented} instruction verification directly into the diffusion and flux training loop.
    

    \item \textbf{Model-Agnostic, \textit{VLM-free} Inferencing:} IABEdit operates as a training-time supervision layer requiring no changes to inference pipelines, making it directly applicable to UNet, transformers in flow-matching architectures with no additional runtime overhead.
    

    \item \textbf{State-of-the-Art Results Across Domains:} IABEdit demonstrates improvements across diverse editing tasks: Scene-level editing and Identity-sensitive Facial editing. IABEdit achieves leading performance on RealEdit, MagicBrush, EMU Edit and D-LORD benchmarks. 
    
    
\end{enumerate}

\subsection{Related Work}

Earlier image manipulation approaches relied on mask-based editing \cite{couairon2023diffedit, singh2024smartmask, xie2023smartbrush}, which provided explicit localized control but required manual spatial annotations. Instruction-based editing methods instead use image generation models \cite{brooks2023instructpix2pix, geng2024instructdiffusion, fu2024guiding, li2025superedit} to embed textual conditions, typically following a triplet setup of source image, instruction, and target edited image. Many earlier works \cite{li2025superedit, gu2025multi} generated training data by rectifying instructions with LLMs and producing edited images via diffusion models. However, current T2I models apply edits in unintended regions, resulting in noisy supervision. To improve quality, \cite{zhang2023magicbrush, fu2024guiding, li2025superedit} introduced high-quality editing datasets through image filtering \cite{hurst2024gpt}. Despite these efforts, most methods still rely on static textual CLIP embeddings \cite{radford2021learning}, used purely as external conditioning. This often leads to misalignment between fine-grained instructions and edited regions. Accurate localization of instruction-relevant elements (e.g., power lines) is crucial for targeted edits (Figure \ref{fig:vis_abstract}\textcolor{black}{b}).

To improve semantic understanding, recent methods \cite{zhang2024hive, gu2025multi} incorporate reward information into text conditions. \cite{huang2024smartedit} replace CLIP with VLMs \cite{liu2023visual, pan2024kosmos} to obtain richer textual understanding. However, these approaches still treat the instruction only as a prior conditioning signal in the diffusion process, without providing active or corrective understandings, limiting fine-grained alignment. They also incur high inference cost due to VLM-based conditioning. SuperEdit \cite{li2025superedit} addresses misalignment through rectified prompts and contrastive supervision, building a discriminative link. Yet, this operates mainly at the prompt level, leaving instruction semantics out of the generation process. Overall, existing methods remain constrained by static conditioning, weak localization, and costly inference, underscoring the need for methods that achieve deeper, more reliable instruction-to-image alignment.

\begin{figure}[t]
    \centering
    \includegraphics[width=\linewidth]{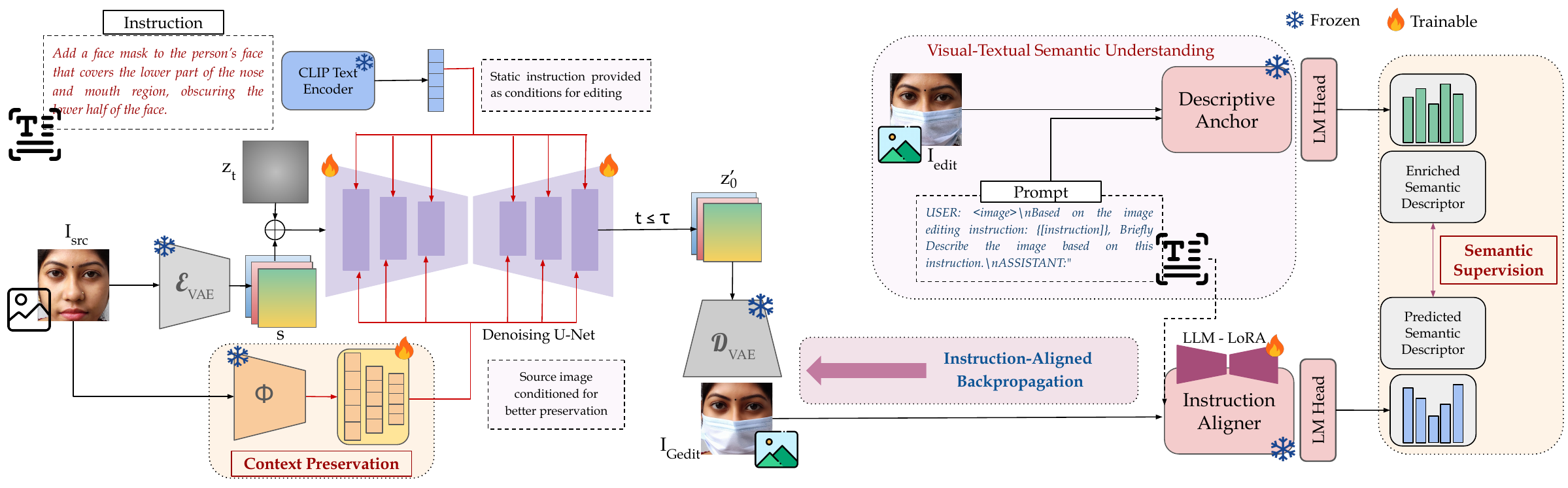}
    \caption{IABEdit enables instruction-based image editing through a semantic-supervised distillation mechanism. A frozen Descriptive Anchor extracts rich guidance from the instruction and ground-truth edit, while an Instruction Aligner learns to align the generated image with this guidance. The alignment is enforced via backpropagation, guiding the diffusion model toward faithful and controllable edits.}
    \label{fig:iabedit_arch}
\end{figure}

\section{IABEdit: Instruction-Aligned Editing}

We propose IABEdit, which provides a balance between context preservation and semantic editing. As illustrated in Figure \ref{fig:iabedit_arch}, our approach consists of two components: (1) image context conditioning, which maintains structural fidelity through encoder and conditioning-level preservation, and (2) editing through gradient-enabled alignment feedback, which provides semantic supervision during training to enforce instruction-consistent modifications.

\subsection{Editing using context as condition}

We formalize successful image editing as a balance between two competing objectives: faithful realization of the instruction and preservation of the original visual context. In practice, many editing models rely primarily on static textual conditioning while using the input image only as a spatial prior, which conditions what needs to be edited but does not focus on what needs to be preserved. Thus, edits may propagate beyond the intended regions or alter unrelated content.

To explicitly enforce context preservation, IABEdit introduces structured conditioning that separates transformation cues from context cues. The generation process is guided by three complementary signals: 
(i) a textual instruction embedding $c_{ins}$ that specifies the desired modification, 
(ii) a source-image embedding $c_{src}$ that encodes high-level visual attributes and context information, and 
(iii) the latent representation $s$ of the source image to retain spatial structure. By decoupling edit intent from contextual cues, this formulation encourages precise modifications while minimizing unintended changes.

 The instruction embedding $c_{ins}$ is obtained from a text encoder, while the source-image embedding $c_{src}$ is derived from visual features of the input image and projected into a conditioning space using a multi-layer linear projector (MLP). These signals are injected into the generative model (e.g., via cross-attention) to guide the editing process. For clarity, we present the diffusion-based formulation in the main paper, while flow-based variants are provided in the \textcolor{blue}{supplementary material}:

\begin{equation}
\label{eq:prop_edit_eq}
\begin{aligned}
c_{ins} &= \text{CLIP}(instruction), \quad 
c_{src} = MLP(\phi(I_{src})), \\
z_{t} &= \sqrt{\bar{\alpha}_t} z_{0} + \sqrt{1 - \bar{\alpha}_t} \, \epsilon, \quad \epsilon \sim \mathcal{N}(0, I), \\
\mathcal{L}_{N} &= \mathbb{E} \left[\left\| \epsilon - \epsilon_\theta \left( z_t \oplus s,\ c,\ t \right) \right\|_2^2\right].
\end{aligned}
\end{equation}

Here, $z_t$ denotes the noisy latent at timestep $t$, $c$ denotes all conditioning inputs, $c=\{c_{\mathrm{ins}},\, c_{\mathrm{src}}\}$. and $\mathcal{L}_N$ is the standard denoising objective. During training, we apply dropout to the instruction embedding $c_{ins}$ and spatial latent $s$ to enable classifier-free guidance. In contrast, the source-image embedding $c_{src}$ is always preserved, as it encodes identity and contextual cues that must remain consistent—particularly in tasks such as face editing.

Importantly, context preservation and edit fulfillment play fundamentally different roles during generation. The contextual cues (s) of the source image remain constant throughout the denoising trajectory, and therefore can be represented as structurally conditioning latents. In contrast, the degree to which the instruction has been satisfied evolves dynamically at each iteration of generation. The regions requiring modification depend on the current state of the image. Static conditioning alone cannot account for this evolving semantic discrepancy. 

This observation motivates the need for a dynamic, output-dependent supervision signal that continuously evaluates how much editing is still required. IABEdit introduces such a mechanism through semantic feedback, enabling the model to adaptively refine edits while preserving context.

\subsection{Semantic Alignment via Gradient-Enabled VLM Supervision}
\label{sec:distill_with_single_step}

\begin{figure*}[t]
    \centering
    \includegraphics[width=0.98\linewidth]{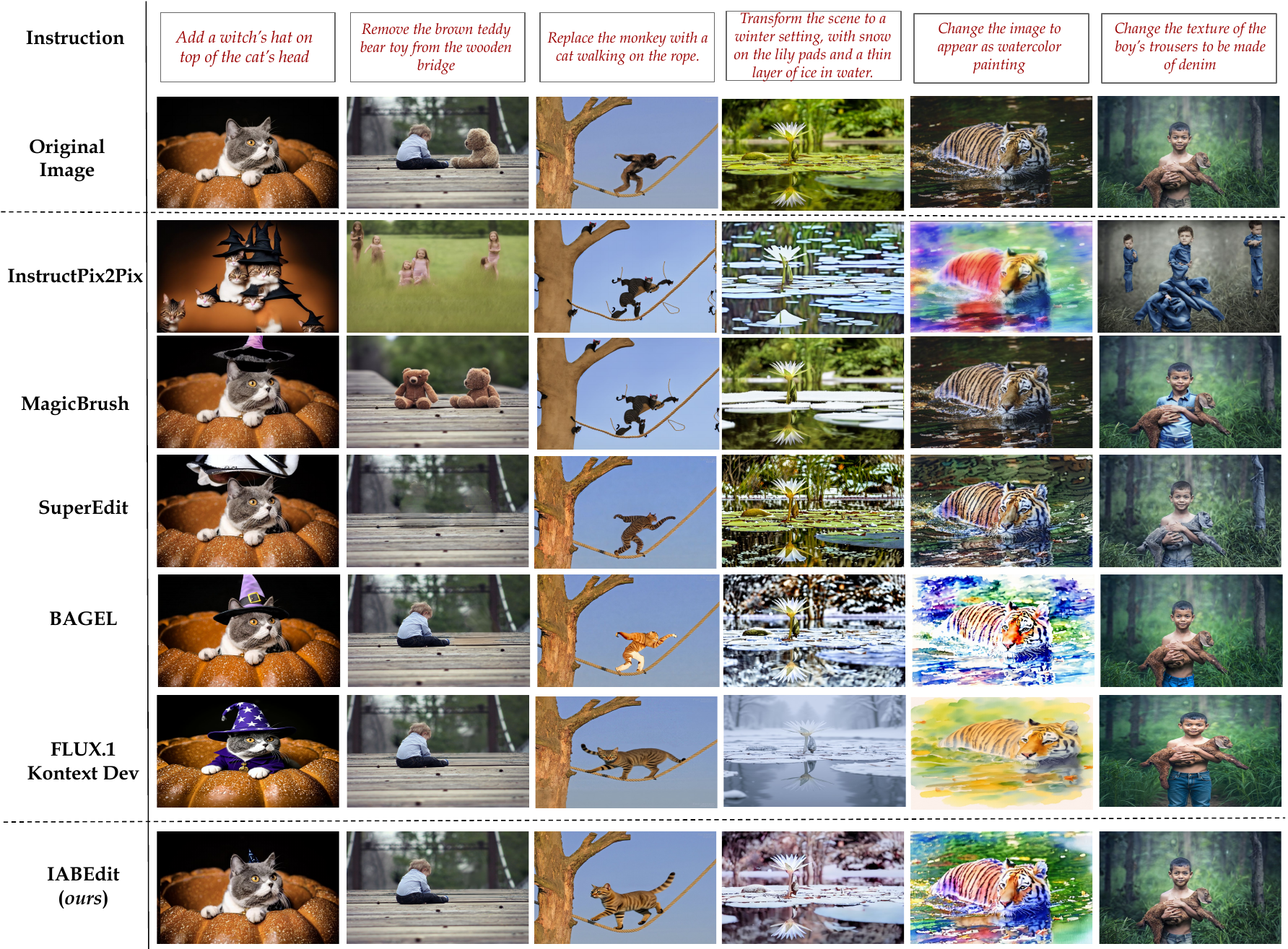}
    \caption{Qualitative results on RealEdit dataset showing comparisons across various editing methods. The instruction adherence can be visualized at the fine-grained level. The non-targeted regions remain preserved, showcasing precise instruction-aligned editing generations. More visual results are available in \textcolor{blue}{supplementary}.}
    \label{fig:qualitative_comparison}
\vspace{-12pt}
\end{figure*}

Conventional instruction-based editing methods encode instructions once via CLIP and inject them as static conditioning signals. This pre-conditioning approach suffers from two critical limitations: short CLIP embeddings lack expressiveness for complex instructions, and the model never verifies whether generated outputs satisfy the instruction. \textcolor{black}{These limitations manifest as systematic failures, as shown visually in Figure \ref{fig:qualitative_comparison}, where \textit{adding a hat "above" rather than "on" the cat's head} indicates an abstract semantic understanding of the instruction}. Further, the image editing \textit{removes both the teddy bear and the child when instructed to remove only the bear}, which shows the lack of spatial localized understanding of the context. Thus, the static conditioning focuses on instruction tokens but lacks a mechanism to infer the underlying intent of the instruction, such as determining edit-remaining regions/tokens.

We address this limitation by leveraging the semantic reasoning capabilities of vision-language models (VLMs). A frozen VLM, Descriptive Anchor, serves as a semantic reference, extracting detailed descriptors from the ground-truth edited image to capture the intended transformation. A trainable VLM, Instruction Aligner, is then optimized to predict these descriptors from the generated output, effectively measuring the discrepancy between the realized and desired edits. The resulting semantic alignment loss produces gradients proportional to the residual discrepancy between the generated image and the intended edit. These gradients are backpropagated to the image generative model, providing output-dependent supervision that adaptively guides both \textit{what} should be modified and \textit{where} modifications should be localized. 

Through this continuous semantic correction process, IABEdit transforms instruction following from static conditioning into iterative alignment, enabling the model to leverage VLM-level understanding during training while learning to perform precise and efficient edits. Importantly, this semantic feedback operates only during training; inference remains a standard generative process without additional computational overhead.

\noindent\textbf{Ground-Truth Semantic Supervision:}
We employ a frozen vision-language model (VLM) to extract rich semantic representations from the ground-truth edited image $I_{edit}$ conditioned on the instruction prompt $P$. Unlike simple captions,
these descriptors encode both global transformations (e.g., “watercolor style applied throughout") and fine-grained localized attributes (e.g., “blue hat on the cat’s head, covering the ears"). This captures spatial relationships, colors, and occlusions far beyond static CLIP embeddings.  As they are automatically generated through image understanding, no additional annotations are required.

\noindent\textbf{Learning Semantic Alignment:}
A trainable VLM, implemented via lightweight LoRA adaptation \cite{hu2022lora}, is optimized to reproduce these semantic representations from the generated image $I_{G_{edit}}$. Matching the generated and reference semantics defines a differentiable alignment objective, enabling gradients to flow back into the image generative model and enforce instruction-consistent edits.

To obtain $z'_0$ without full diffusion sampling, we employ single-step denoising at small timesteps $t \leq \tau$. At these early timesteps, noisy latent $z'_t$ closely approximates clean latent $z_0$ since noise level $\sqrt{1 - \bar{\alpha}_t}$ is minimal:

\begin{equation}
\label{eq:ctr_pp_eq}
z_0 \approx z'_0 = \frac{z'_t - \sqrt{1 - \bar\alpha_t} \, \epsilon_\theta((z'_t \oplus s), c_{ins}, c_{src}, t)}{\sqrt{\bar\alpha_t}}
\end{equation}

This approximation~\cite{controlnet_plus_plus} provides precise estimates while avoiding the memory overhead of 50+ DDIM steps per training batch.

\noindent\textbf{Alignment-Relevant gradient flow:}
We train our dynamic semantic aligner via KL divergence minimization:
\begin{equation}
\begin{aligned}
\label{eq:distill_eq}
\mathcal{L}_{\text{D}} &= \frac{1}{B} \sum_{b=1}^{B} \text{KL} \left( 
\text{Softmax} \left( \frac{\mathbf{l}_f^{(b)}}{T} \right) \Bigg\| 
\text{Softmax} \left( \frac{\mathbf{l}_t^{(b)}}{T} \right) 
\right) * T^2 \\
&= \frac{1}{B} \sum_{b=1}^{B} \sum_{i=1}^{N} p_f^{(b)}(i) 
\left[ \log p_f^{(b)}(i) - \log p_t^{(b)}(i) \right] * T^2
\end{aligned}
\end{equation}
where $\mathbf{l}_f^{(b)}$ and $\mathbf{l}_t^{(b)}$ denote the token-level logits from the frozen Descriptive Anchor and the trainable semantic aligner for the $b$-th sample over $N$ tokens. A temperature $T$ is used to soften the probability distributions, enabling the aligner to learn from fine-grained semantic discrepancies; the $T^2$ factor compensates for gradient scaling during distillation. The corresponding token probabilities are denoted as $p_f^{(b)}(i)$ and $p_t^{(b)}(i)$.

Crucially, spatial localization emerges through gradient flow. The alignment loss $\mathcal{L}_{\text{D}}$ is computed from the generated image $I_{G_{edit}}$, which depends on the predicted latent $z'_0$ and ultimately on the model’s noise estimator $\epsilon_\theta$. Backpropagating $\nabla \mathcal{L}_{\text{D}}$ therefore injects semantic discrepancy signals directly into the generative backbone, guiding updates toward regions that require modification to satisfy the instruction. Unlike static conditioning, which provides fixed semantic priors, IABEdit embeds semantic reasoning verification into the optimization objective itself. Instruction adherence is no longer assumed—it is explicitly minimized as part of the training landscape.

\noindent\textbf{Joint Optimization of Context Preservation:}
Distillation gradients also update the MLP projector mapping $\phi(I_{src})$ to source embedding $c_{src}$. Rather than encoding generic features, the projector learns instruction-aware conditioning: emphasizing features in regions requiring edits while preserving unchanged areas. This joint optimization of textual (via distillation) and visual (via MLP) conditioning enables spatially precise edits.

\noindent This framework enables faithful instruction-guided editing across general scenes; however, certain domains such as face editing require additional constraints to preserve subject-specific identity characteristics during modification.
\newline

\noindent\textbf{Identity Preservation for Face Editing: } For face editing applications in surveillance and security contexts, maintaining biometric identity under occlusions (masks, sunglasses) is critical. While semantic distillation ensures instruction alignment, it does not explicitly constrain identity preservation. We therefore introduce an identity coherence loss $\mathcal{L}_{C}$ that minimizes cosine distance between facial embeddings:
\begin{equation}
\label{eq:cons_eq}
\mathcal{L}_{\text{C}} = \frac{1}{B} \sum_{b=1}^{B} \left(1 - \cos(\phi^{(b)}(I_{src}), \phi^{(b)}(\mathcal{D}_{\text{VAE}}(z'_{0})))\right)
\end{equation}

\noindent where $\phi$ is a pre-trained face recognition model (ArcFace \cite{deng2019arcface}). This encourages retention of biometric features in the generated edit $\mathcal{D}_{\text{VAE}}(z'_0)$ despite occlusions or appearance changes. By operating in the embedding space of a discriminatively trained face recognition model, this loss ensures that identity-critical features (facial structure, proportions) remain invariant even when instruction-guided edits add realistic occlusions that would otherwise distort facial geometry.

The complete loss function combines all components conditionally:
\begin{equation}
\label{eq:iab_edit_eq}
\mathcal{L}_{\text{IABEdit}} =
\left\{
\begin{aligned}
\lambda_{N} \cdot \mathcal{L}_{N} + \lambda_{D} \cdot \mathcal{L}_{D} + \lambda_{C} \cdot \mathcal{L}_{C}, & \quad \text{if } t \leq \tau \\
\lambda_{N} \cdot \mathcal{L}_{N}, & \quad \text{otherwise}
\end{aligned}
\right.
\end{equation}

\noindent where \( \lambda_{\text{N}} \), \( \lambda_{\text{D}} \), and \( \lambda_{\text{C}} \) are scalar weights that balance the contributions of each loss term. $\tau$ refers to the timestep threshold, which is a hyper-parameter used to determine whether the generated edited image should be utilized for VLM fine-tuning and maintaining structural coherence of faces. Distillation ($\mathcal{L}_D$) and identity ($\mathcal{L}_C$) losses apply only at small timesteps ($t \leq \tau$) where generated outputs are available via single-step denoising. For general scene editing tasks, $\mathcal{L}_C$ is omitted, and the framework optimizes only $\mathcal{L}_N$ and $\mathcal{L}_D$.

\section{Experimental Details}

We present a detailed experimental setup for rigorously evaluating our editing mechanism with various benchmarks, a model-agnostic generative method and balanced evaluation criteria. 

\noindent\textbf{Datasets:} For scene image editing tasks, we use Super-40K \cite{li2025superedit} as a training dataset. Evaluation is conducted on RealEdit \cite{gu2025multi}, the MagicBrush \cite{zhang2023magicbrush}, and the EMU Edit \cite{sheynin2024emu} test set to assess image preservation and instruction alignment. To assess the performance of the model in identity-sensitive applications such as surveillance, we further train and evaluate on the disjoint train-test sets of D-LORD \cite{manchanda2023d} dataset \footnote{The dataset was obtained through: \url{https://dyslai.org/}}, which includes facial images under occlusion.

\noindent\textbf{Architecture and Implementation Details:} The IABEdit framework is model-agnostic, can work with diffusion and flow-based generative models. Semantic supervision is provided using a LLaVA-based \cite{liu2023visual} VLM model. During inference, IABEdit functions as a lightweight diffusion-only model. It uses CLIP-based 77-token instruction embeddings (where instructions comprise 20-30 effective textual tokens) and conditional image features (CLIP for scene edits and ArcFace for facial edits) with a learned small set of MLP layers. To incorporate deeper semantic guidance with detailed descriptive context, we use LLaVA-based 1362-token (with 100 effective generated textual tokens) descriptors, enabling fine-grained edits aligned with instruction semantics. Additional implementation details, and training and testing protocols are provided in the \textcolor{blue}{supplementary}.

\noindent\textbf{Evaluation Metrics (Alignment and Preservation):} IABEdit is evaluated in terms of quantitative metrics, agent-based assessments, as well as human evaluations. To assess localized edits and structural preservation, we use feature-level metrics: \textcolor{black}{CS-P (Content Similarity Preserving) for measuring content retention of non-edited regions by computing CLIP-similarity between the source and edited image, CLIP-I for measuring similarity to the target-edited image, and CLIP-T for instruction alignment that leverages a pre-trained CLIP model.} DINO-I \cite{caron2021emerging} and L1 scores measure structural and pixel-level similarity to the ground truth edited image, while DINO-P is used to measure the facial identity structural preservation of the edited face with respect to source face. To capture the trade-off between instruction adherence and content preservation, we compute Harmonic Mean (HM) between CLIP-T and CS-P, offering a unified measure of balanced editing performance. This helps quantify whether an edit faithfully follows the instruction without significantly altering the non-editing regions. As traditional metrics often miss perceptual and semantic subtleties that humans easily identify \cite{huang2024smartedit, li2025superedit, gu2025multi}, we incorporate MLLM agents like GPT-4o \cite{hurst2024gpt} to evaluate alignment, spatial precision, and realism, complemented by a user study grounded in the same criteria.

\section{Results and Analysis}

We evaluate IABEdit across multiple datasets and compare it with state-of-the-art instruction-based editing methods. 
While recent work has improved performance through dataset curation \cite{hui2025hqedit, zhang2023magicbrush} and instruction refinement \cite{li2025superedit, gu2025multi}, a fundamental challenge remains: directly aligning textual semantics with precise spatial editing regions during training. As shown in Figure \ref{fig:qualitative_comparison}, IABEdit addresses this challenge through gradient-based semantic distillation, yielding superior instruction adherence and localization.

\noindent\textbf{Instruction Alignment and Localized Edit Control:} As shown in Figure \ref{fig:qualitative_comparison}, existing methods struggle with precise localization. For instance, SuperEdit removes not only the teddy bear but also the child (column 2), and adds denim texture to both the boy's trousers and the goat (column 5). Whereas an autoregressive multi-modal generative method, such as BAGEL, also fails to completely remove the teddy bear from the bridge \textit{(zoom in for a better view)}, and also just changes the color of the trousers without changing their texture to denim. Similarly, a lack of understanding of the instructions' semantics is observed in recent FLUX.1 Kontext Dev method, it over-edits as it not only adds the hat on the cat's head, but it also puts a jacket on the cat (see Figure \ref{fig:qualitative_comparison} column 1), similarly, it changes the trousers to denim but also increases the length of the boy's legs. These errors reflect weak coupling between instruction semantics and spatial edits. In contrast, IABEdit's gradient-based distillation enables fine-grained control, modifying only instruction-relevant regions. Quantitatively also, as shown in Tables \ref{tab:real_edit_quant_tab} and \ref{tab:magic_brush_quant_tab}, IABEdit with diffusion achieves the highest CLIP-T scores among diffusion-based methods on RealEdit (27.60) and MagicBrush (30.97) benchmarks, which is superior to recent diffusion models (UltraEdit and SuperEdit) and perform competitive to computationally heavy flux-based (FLUX.1 Kontext Dev) and autoregressive (BAGEL) image editing methods. IABEdit also achieved a CLIP-I score of 92.41 on the MagicBrush test set, indicating strong alignment between the edited image and the target edited image, showing the controlled localized edit.

As shown in Table \ref{tab:magic_brush_quant_tab}, IABEdit achieves the best DINO-I score on MagicBrush (88.26), outperforming the strongest \textcolor{black}{diffusion-based baseline (UltraEdit) by +3.49 points}, and attains the lowest L1 distance (0.058), indicating sharper structural fidelity to the ground-truth edits. Moreover, GPT-4o score evaluations (Table \ref{tab:real_edit_agent_tab} "Following" dimension) also rank IABEdit highly for instruction adherence (3.63), affirming its semantic precision in fine-grained edits.

\begin{table}[t]
\centering
\caption{ (a) Comparing performance of IABEdit (Diffusion+LLaVA) with other image editing methods on the RealEdit dataset. (b) Comparing performance of IABEdit (Diffusion+LLaVA) with other image editing methods on the MagicBrush benchmark.}
\vspace{-18pt}
\begin{subtable}{0.49\linewidth}
\centering

\caption{\label{tab:real_edit_quant_tab}} 
\vspace{-6pt}
\resizebox{0.82\linewidth}{!}{
\begin{tabular}{l|c|c|c}
\hline
\textbf{Method} & \textbf{CS-P} $\uparrow$ & \textbf{CLIP-T} $\uparrow$ & \textbf{HM} $\uparrow$ \\ \hline
MagicBrush \cite{zhang2023magicbrush} & 90.25 & 26.98 & 41.28\\
InstructDiffusion \cite{geng2024instructdiffusion} & 74.20 & 26.69 & 38.71 \\
InstructPix2Pix \cite{brooks2023instructpix2pix} & 75.76 & 27.00 & 39.78 \\
Reward-InstructPix2Pix \cite{gu2025multi} & 89.69 & 27.31 & \underline{41.67} \\
MGIE \cite{fu2024guiding} & 84.64 & 26.20 &  39.58 \\
SmartEdit \cite{huang2024smartedit} & 87.70 & 26.50 & 40.64 \\
\textcolor{black}{HQ-Edit} \cite{hui2025hqedit} & 66.32 & 27.09 &  38.21 \\
\textcolor{black}{UltraEdit} \cite{zhao2024ultraedit} & 84.91 & \underline{27.50} & 41.39 \\
SuperEdit \cite{li2025superedit} & \textbf{90.93} & 26.01  & 40.24 \\
\textit{BAGEL} \cite{deng2025emerging} & \underline{90.60} & 26.69 & 41.00 \\
\textit{FLUX.1 Kontext Dev} \cite{labs2025flux} & 87.64 & 27.21 & 41.29 \\
\rowcolor{lightred}
\textit{IABEdit (Ours)} & 89.65 & \textbf{27.60} & \textbf{42.00} \\
\hline
\end{tabular}}
\end{subtable}
\hfill
\begin{subtable}{0.49\linewidth}
\caption{\label{tab:magic_brush_quant_tab} } 
\vspace{-6pt}
\resizebox{\linewidth}{!}
{
\begin{tabular}{l|c|c|c|c}
\hline
\textbf{Method} & \textbf{CLIP-I} $\uparrow$ & \textbf{CLIP-T} $\uparrow$ & \textbf{DINO-I} $\uparrow$ & \textbf{L1} $\downarrow$ \\ \hline
MagicBrush \cite{zhang2023magicbrush}   & 90.70 & 30.60 & 80.60 & \underline{0.062} \\
InstructDiffusion \cite{geng2024instructdiffusion}   & 89.20 & 30.20 & 77.70 & - \\
InstructPix2Pix \cite{brooks2023instructpix2pix}   & 85.40 & 29.20 & 69.80 & 0.112 \\
Reward-InstructPix2Pix \cite{gu2025multi} & 88.90 & 29.80 & - & - \\
MGIE \cite{fu2024guiding} & 90.90 & 30.50 & - & - \\
SmartEdit \cite{huang2024smartedit}   & 90.40 & 30.30 & 79.70 & 0.081 \\
\textcolor{black}{HQ-Edit} \cite{hui2025hqedit} & 69.08 & 25.88 &  51.62 & 0.243 \\
\textcolor{black}{UltraEdit} \cite{zhao2024ultraedit} & 90.47 & 30.86 & 84.77 & 0.066\\
SuperEdit \cite{li2025superedit}   & \multicolumn{1}{c|}{90.50} & 30.30 & 80.20 & 0.106 \\ 
\textit{BAGEL} \cite{deng2025emerging} & \underline{92.13} & \underline{31.02} & \underline{87.00} & - \\
\textit{FLUX.1 Kontext Dev} \cite{labs2025flux} & 91.90 & \textbf{31.28} & 86.03 & -\\
\rowcolor{lightred}
\textit{IABEdit (Ours)} & \textbf{92.41} & 30.97 & \textbf{88.26} & \textbf{0.058} \\
\hline
\end{tabular}}
\end{subtable}
\vspace{-12pt}
\end{table}

\noindent\textbf{Content and Structural Preservation:} IABEdit excels at maintaining the integrity of non-instructed image regions, preserving both visual content and spatial structure.
It can be observed from Figure \ref{fig:qualitative_comparison}, that recent state-of-the-art FLUX-based editing method (FLUX.1 Kontext Dev) does not preserve the context of the image as shown through the distortion in the background context behind the lily flowers (column 3, row 6), and in the `watercolor-style edit’, it fails to preserve the water region present in the original tiger image. Quantitatively, this is reflected by achieving high DINO-P and CS-P scores across datasets, indicating that the original image semantics are retained even after editing. Moreover, IABEdit achieves superior DINO-P scores, particularly on the D-LORD dataset, demonstrating its strength in preserving identity-specific structural features, especially in face editing tasks as compared to a proprietary image-generation Gemini-AI agent. The framework’s use of cross-attention and latent conditioning enables localized edits without unintended modifications. Together, these results highlight IABEdit’s ability to perform precise edits while preserving the non-edited regions of original image.

\begin{table}[t]
\centering
\caption{\label{tab:real_edit_agent_tab} Performance comparison on the RealEdit dataset (having a sample size of 560 instruction-image pairs) across various instruction-based image editing methods. The evaluations are carried out using GPT-4o.}
\resizebox{0.85\linewidth}{!}{
\begin{tabular}{l|cc|cc|cc|cc}
\hline
\multirow{2}{*}{\textbf{Method}} & \multicolumn{2}{c|}{\textbf{Following} $\uparrow$} & \multicolumn{2}{c|}{\textbf{Preserving} $\uparrow$} & \multicolumn{2}{c|}{\textbf{Quality} $\uparrow$} & \multicolumn{2}{c}{\textbf{Overall} $\uparrow$} \\ \cline{2-9} 
& 
\textbf{Acc (\%)} & \textbf{Score} & \textbf{Acc (\%)} & \textbf{Score} & \textbf{Acc (\%)} & \textbf{Score} & \textbf{Acc (\%)} & \textbf{Score} \\ \hline
MagicBrush \cite{zhang2023magicbrush}   & 51.00 & 2.90 & 70.00 & 3.85 & 50.00 & 3.67 & 57.00 & 3.47 \\
InstructDiffusion \cite{geng2024instructdiffusion}   & 52.00 & 2.87 & 54.00 & 3.17 & 45.00 & 3.58 & 50.30 & 3.21 \\
InstructPix2Pix \cite{brooks2023instructpix2pix}   & 52.00 & 2.94 & 53.00 & 3.31 & 50.00 & 3.69 & 51.70 & 3.31 \\
Reward-InstructPix2Pix \cite{gu2025multi}   & 63.00 & 3.39 & 58.00 & 3.43 & 54.00 & 3.80 & 58.30 & 3.54 \\
MGIE \cite{fu2024guiding}   & 40.00 & 2.43 & 45.00 & 2.79 & 38.00 & 3.35 & 41.00 & 2.86 \\
SmartEdit \cite{huang2024smartedit}   & 64.00 & 3.50 & 66.00 & 3.70 & 45.00 & 3.56 & 58.30 & 3.59 \\
SuperEdit \cite{li2025superedit}   & \textbf{67.00} & \underline{3.59} & \textbf{77.00} & 4.14 & \textbf{65.00} & \textbf{4.01} & \textbf{69.70} & \textbf{3.91} \\ 
\rowcolor{lightred}
\textit{IABEdit (Ours)} &  \underline{66.00} & \textbf{3.63} & \underline{72.00} & \textbf{4.19} & \underline{54.00} & 3.54 & \underline{64.00} & \underline{3.78} \\ \hline
\end{tabular}}
\end{table}

\noindent\textbf{Balanced Editing: Preserving While Transforming:} IABEdit optimally balances instruction adherence and content preservation, achieving the highest Harmonic Mean (HM) between CLIP-T and CS-P metrics (Table \ref{tab:real_edit_quant_tab}). On RealEdit, IABEdit scores HM=42.00, significantly outperforming Reward InstructPix2Pix (41.67), \textcolor{black}{UltraEdit (41.39)}, SuperEdit (40.24), BAGEL (41.00), FLUX.1 Kontext Dev (41.29) and other baselines. We also performed Wilcoxon signed-rank test with n=560 and observed significant results for all models with $p<$ 0.05, \textcolor{black}{whose detailed values are shown in the} \textcolor{blue}{supplementary}. 

IABEdit's CS-P score of 89.65 is slightly lower than SuperEdit's 90.93, but this difference reflects successful editing rather than a limitation. When instructions require substantial changes, the edited image naturally diverges from the source. Many competing methods struggle with this balance: they either preserve too much content and under-edit (high CS-P but low CLIP-T), or they over-edit and introduce unintended changes (low CS-P with inconsistent CLIP-T). IABEdit's gradient-based distillation provides precise spatial control, modifying only the regions specified by the instruction while leaving other areas unchanged.

\noindent\textbf{Agent-Based and Human-Centric Evaluation:} 
\textcolor{black}{Table \ref{tab:real_edit_agent_tab} reports GPT-4o evaluations on RealEdit across instruction following, content preservation, and image quality measures. \textit{Score} measures fine-grained assessment on a 0-5 scale while \textit{Accuracy} acts as a binary acceptability criterion. Although SuperEdit is more consistent at passing the binary baseline, IABEdit attains higher continuous scores in Instruction Following (3.63) and Content Preservation (4.19), indicating more precise edits while retaining non-target regions. IABEdit's perceptual quality score (3.54) is slightly lower than SuperEdit, but the overall score remains competitive at 3.78 (second-best). The quality-based comparisons are discussed in the \textcolor{blue}{supplementary} material. Overall, it reveals a trade-off: SuperEdit offers robustness, while IABEdit provides higher-fidelity edits and preservation. On D-LORD facial edits, IABEdit attains an instruction-following score of 4.67 with 94.8\% accuracy, comparable to Gemini-AI (4.53, 92.19\%).}

\begin{figure}[t!]
    \centering
    \includegraphics[width=0.85\linewidth]{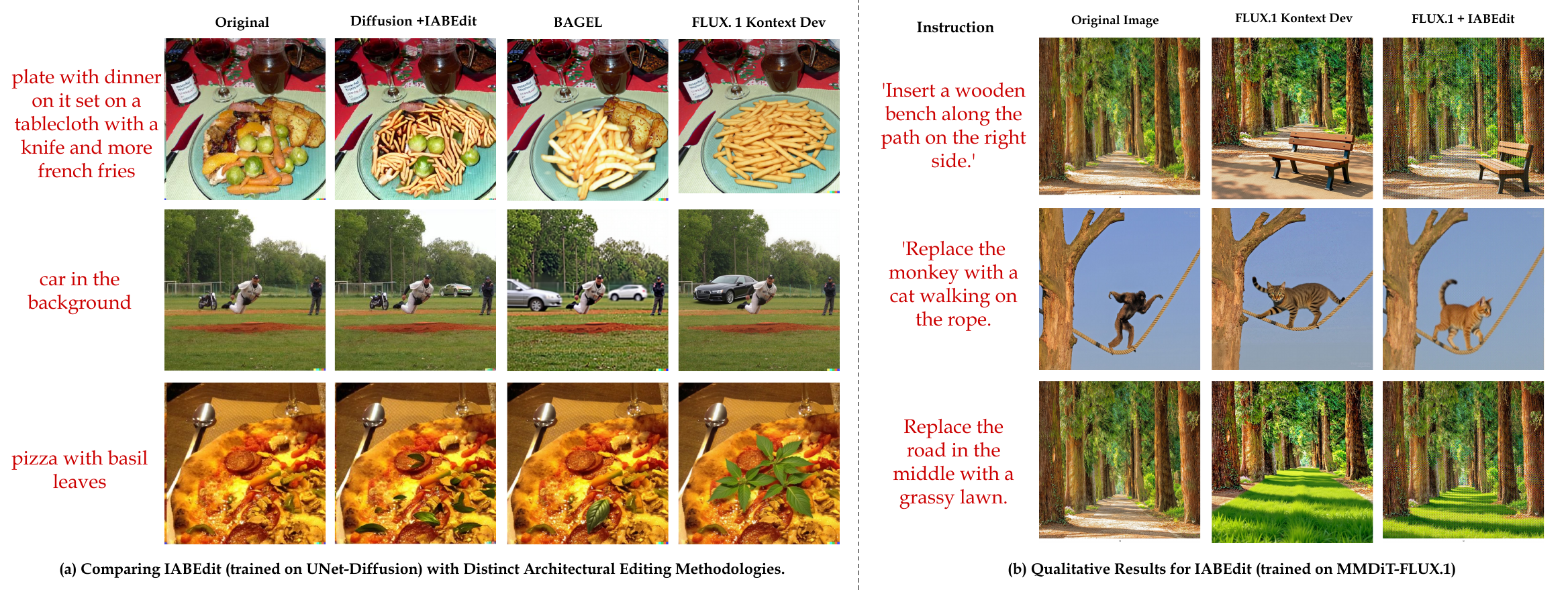}
    \caption{(a) Qualitative comparison of IABEdit (+UNet Diffusion) on MagicBrush test-set with other state-of-the-art editing methods. (b) Qualitative visualization of RealEdit test-set samples showcasing improvement of FLUX.1 method by training it with the IABEdit framework.}
    \label{fig:editing_methods}
    \vspace{-22pt}
\end{figure}

\noindent\textbf{Comparison with Distinct Architectural Editing Methodologies:} We compare IABEdit against fundamentally different architectures: FLUX.1 Kontext Dev\cite{labs2025flux} (flow matching) and BAGEL \cite{deng2025emerging} (unified autoregressive multi-modal mixture of transformers). Although these methods generate photorealistic output, they exhibit three failure modes (Figure \ref{fig:editing_methods}\textcolor{black}{a}): \textit{Keyword fixation:} focusing on isolated words ("fries", "car") rather than full instruction semantics. \textit{Spatial misalignment:} inserting objects without respecting scene composition (e.g., overlaying a car on an existing bike).
\textit{Global distortion:} altering color/lighting beyond edited regions, and overshooting scale by adding "too large basil leaves" without respecting the image content's scale.

We observe that, even with a lightweight UNet-based diffusion method in the IABEdit framework, it performs better-balanced image editing (refer to Figure \ref{fig:editing_methods}(a)--qualitative visualization and quantitative performance observed through Table \ref{tab:real_edit_quant_tab} and Table \ref{tab:magic_brush_quant_tab}) as compared to the computationally heavy flux and autoregressive methods.

\noindent\textbf{Model Agnostic IABEdit Framework:} Table \ref{tab:real_edit_quant_tab} and \ref{tab:magic_brush_quant_tab} show results with IABEdit using a UNet-based denoising backbone, which outperforms existing diffusion-based methods such as SuperEdit, InstructPix2Pix and Reward InstructPix2Pix. Moreover, when IABEdit is trained over FLUX.1 model, it improves the image quality of the generated edited image, this property is inherited from FLUX's quality. As we noted that FLUX.1 is instruction's keyword fixated and does not understand the instruction's semantics, further it also does not focus on the image's context. We solve this problem by training FLUX.1 with the IABEdit framework, which improves its image-text semantic alignment and also preserves the image context in the generated edited image. After training FLUX.1 with IABEdit, it improves FLUX.1 on the RealEdit dataset by attaining a CS-P score of \textit{\textbf{90.20}}, CLIP-T score of \textit{\textbf{27.78}} and performs balanced editing by achieving a HM score of \textit{\textbf{42.28}}. While FLUX.1 achieves a CS-P score of \textit{\textbf{87.64}}, a CLIP-T score of \textit{\textbf{27.21}} and performs balanced editing by achieving a HM score of \textit{\textbf{41.29}}. All these scores achieved on IABEdit+FLUX.1 are significantly higher, as they attained a p-value<0.05. It can be observed that IABEdit+FLUX.1 improves the image context preservation score compared to the existing FLUX.1 model. Figure \ref{fig:editing_methods}(b) presents qualitative results of FLUX.1+IABEdit. It can be observed that the context of the original image is preserved in IABEdit (+FLUX.1) while following the editing instruction, whereas FLUX.1 (Kontext Dev) destroys the image context. Therefore, IABEdit is a model-agnostic framework suitable for all kinds of denoising networks, including UNet in Stable Diffusion and MMDiTs in FLUX.1.

\textcolor{black}{Results on the challenging EMU Edit benchmark \cite{sheynin2024emu} across diffusion and FLUX-based architectures are provided in the \textcolor{blue}{supplementary}. IABEdit outperforms the strongest baseline by +1.41 in CLIP-based instruction following, +0.84 in target-caption alignment, and +9.98 in DINO-based structural preservation, demonstrating semantically accurate edits while preserving image structure.}

\begin{wraptable}{r}{0.51\linewidth}
\centering
\vspace{-32pt}
\caption{\label{tab:dlord_quant_tab} Quantitative comparison with agent-based generation models on the real-world D-LORD dataset for facial editing.}
\resizebox{\linewidth}{!}{
\begin{tabular}{l|c|c|c|c}
\hline
\textbf{Generation Technique} & \textbf{CS-P} $\uparrow$ & \textbf{CLIP-T} $\uparrow$ & \textbf{HM} $\uparrow$ & \textbf{DINO-P} $\uparrow$ \\ \hline
SuperEdit \cite{li2025superedit} & \underline{79.19} & 22.96 & 35.38 & \underline{51.64} \\
Gemini-AI (Agent) \cite{comanici2025gemini} & 78.67 & \underline{31.40} & \underline{44.79} & 50.20 \\
\rowcolor{lightred} 
\textit{IABEdit (Ours)} & \textbf{79.93} & \textbf{33.10} & \textbf{46.74} & \textbf{55.33} \\ 
\hline
\end{tabular}}
\vspace{-22pt}
\end{wraptable}

\noindent\textbf{Real-World Surveillance Scenarios:} In real-world surveillance scenarios, instruction-based facial editing is particularly challenging due to realistic transformations like disguises (masks and sunglasses), while preserving identity. As shown in Table \ref{tab:dlord_quant_tab}, IABEdit outperforms the Gemini-AI (2.5Pro) \cite{comanici2025gemini} \textcolor{black}{and SuperEdit methods}, achieving higher CS-P (79.93) and CLIP-T (33.10) scores. We also performed a Wilcoxon signed-rank test on the harmonic mean scores and achieved a p-value of $<$ 0.05. An improvement of +5.13 DINO-P points over \textcolor{black}{Gemini-AI (agent) method and +3.69 over SuperEdit (trained on D-LORD train set and evaluated on disjoint identity test set)} indicates stronger preservation of structural identity. We have also evaluated the effectiveness in terms of face recognition performance. While the detailed results are included in \textcolor{blue}{supplementary}, combining the generated edited facial samples with the real face dataset improved the recognition performance by 2.82\%.

\begin{figure}[t]
\centering

\begin{minipage}[t]{0.33\textwidth}
     \caption{Ablation results on the Real-Edit dataset. }
     \vspace{-9pt}
     \includegraphics[width=\linewidth]{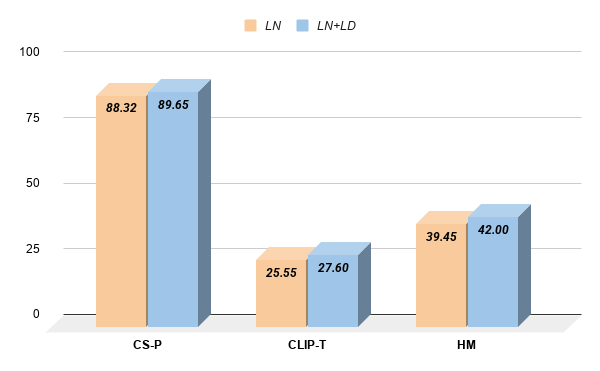}

\label{fig:ablation_realedit_tab}
\end{minipage}
\hfill
\begin{minipage}[t]{0.6\textwidth}
    \centering
\vspace{-12pt}
\captionof{table}{\label{tab:flux_rebuttal} Comparisons on RealEdit dataset using 3 protocols: (1) with and w/o LoRA-finetuning, (2) VLM models, and (3) with and w/o context preservation conditioning.}
\vspace{5pt}
\resizebox{\linewidth}{!}{
\begin{tabular}{l|c|c|c}
\hline
 & \textbf{CS-P ($\uparrow$)} & \textbf{CLIP-T ($\uparrow$)} & \textbf{HM ($\uparrow$)} \\ \hline
Without LoRA-finetuning & 83.44 & 24.44 & 37.58 \\
\rowcolor{lightred} 
\textbf{With LoRA-finetuning (IABEdit)} & \textbf{89.65} & \textbf{27.60} & \textbf{42.00} \\ \hline
IABEdit (LLaVA-7B) & 89.65 & 27.60 & 42.00 \\ 
\rowcolor{lightred} 
\textbf{IABEdit (Qwen-VL 2.5-7B)} & 89.33 & \textbf{27.91} & \textbf{42.27} \\ \hline
IABEdit (w/o context preservation) & 88.91 & 25.90 & 40.19 \\
\rowcolor{lightred} 
\textbf{IABEdit (w/ context preservation)} & \textbf{89.65} & \textbf{27.60} & \textbf{42.00} \\ \hline
\end{tabular}}
\end{minipage}

\end{figure}

\noindent\textbf{Ablation Study:} \noindent\textit{\textbf{(1) IABEdit's Component Ablation:}}  Figure \ref{fig:ablation_realedit_tab} presents the ablation results highlighting the direct impact of the proposed IABEdit method based on the loss components introduced in our method. The effectiveness of our method for balanced editing is demonstrated by higher HM scores achieved when training the model with combined Denoising ($\mathcal{L}_{N}$) and Distillation ($\mathcal{L}_{D}$) loss functions. Incorporating both objectives results in +2.47 performance improvement compared to training with only Denoising loss ($\mathcal{L}_{N}$). This underscores the importance of the distillation component with gradient-enabled meaningful supervisory signals. It enhances the image generation model’s ability to perform semantically aligned edits with natural language instructions.

\noindent \textit{\textbf{(2) Impact of LoRA fine-tuning:}} Table \ref{tab:flux_rebuttal} (first two rows) shows ablation results for highlighting the utilization of LoRA adapter for fine-tuning the predicted semantic VLM descriptor. Without LoRA adaptation, the Instruction Aligner operates on noisy intermediate denoised outputs and produces unstable semantic descriptors, resulting in inaccurate gradient updates from $\nabla \mathcal{L}_D$. LoRA adaptation significantly improves robustness to intermediate generations, enabling semantic supervision, while the frozen supervisor is a stable reference. This "clean expert + noise-robust learner" design outperforms using a single frozen VLM for both roles. LoRA-based finetuning results in balanced image editing, with a large improvement of \textbf{+4.42} HM scores compared to w/o LoRA.

\noindent\textit{\textbf{(3) Stronger VLM backbone:}} To study the impact of different VLMs, we replaced LLaVA-7B with Qwen-VL-2.5-7B \cite{bai2025qwen25vltechnicalreport}, later further improves alignment (CLIP-T: 27.60 $\rightarrow$ 27.91) and balanced editing (HM: 42.00 $\rightarrow$ 42.27), indicating that stronger semantic reasoning translates to better guidance.

\noindent\textcolor{black}{\noindent\textit{\textbf{(4) Context-preservation conditioning:}} Table \ref{tab:flux_rebuttal}(last two rows) present results with and without context-preservation conditioning to highlight its effectiveness. With conditioning, CS-P improves by +0.74, CLIP-T by +1.70, and HM by +1.81, demonstrating more balanced image editing.}

\noindent\textbf{Human Study:} We have conducted a user study to understand the human-centric evaluations on the image editing task. The user study is conducted on the challenging RealEdit dataset with $30$ subjects aged $20-45$. Evaluations are carried out across three dimensions: Instruction following, Non-instructed region preservation and Quality of generated edited image. Each participant selected the best-performing model in each evaluation metric case for the randomly provided $30$ triplets of generated edited image, original image and the editing instruction. The triplets are provided for multiple instruction-based models such as InstructPix2Pix \cite{brooks2023instructpix2pix}, MagicBrush \cite{zhang2023magicbrush}, SuperEdit \cite{li2025superedit} and the proposed IABEdit. 

\begin{wrapfigure}{r}{0.47\linewidth}
  \begin{center}
    \includegraphics[width=\linewidth]{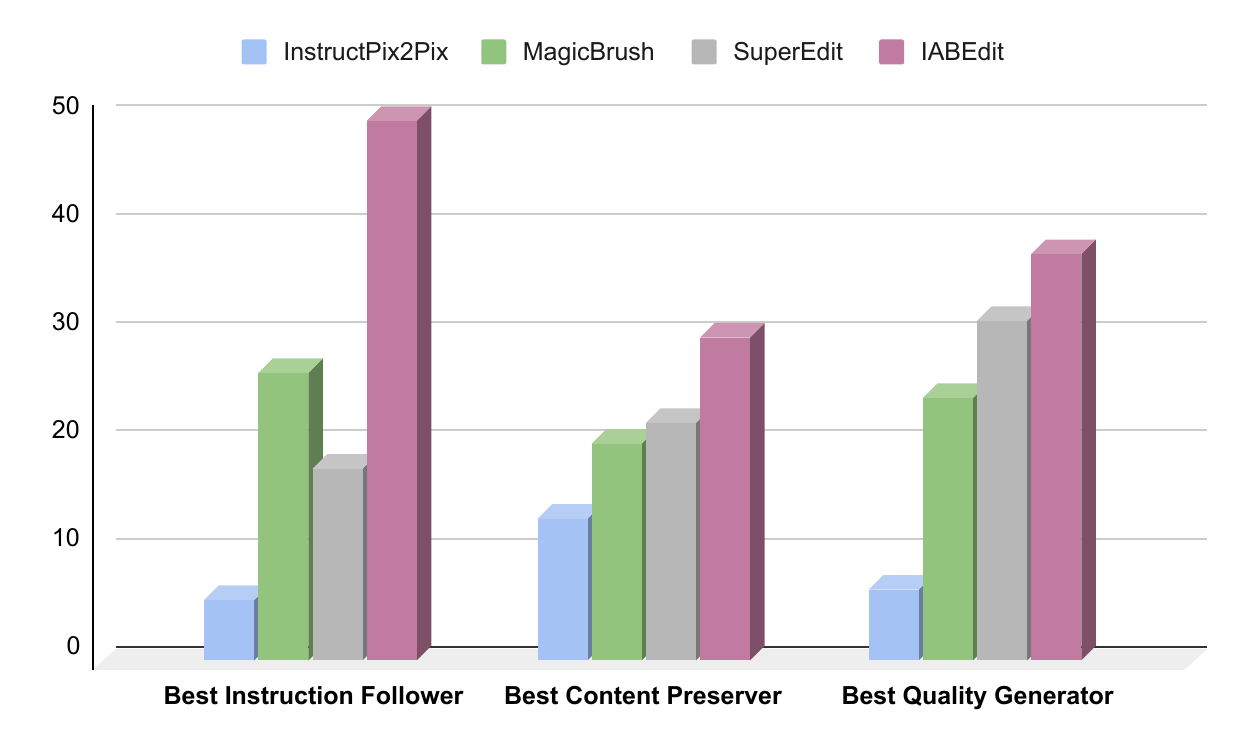}
\vspace{-30pt}
  \end{center}
  \caption{User study results on RealEdit dataset. The values indicate the best performing model in randomly selected 30 images. Three metrics for human evaluations: Instruction Following, Non-Editing Content Preserving and Quality of Generated Edited image. }
  \label{fig:user_study_plot}
  \vspace{-27pt}
\end{wrapfigure}
Figure \ref{fig:user_study_plot} shows the user responses for the best-performing model on $30$ randomly selected images across all four instruction-based models. IABEdit achieves the highest Instruction Following rate of \textbf{50.00\%}, demonstrating more accurate adherence to editing instructions than the competing methods. It also best preserves the non-instructed regions, achieving the highest preservation rate of \textbf{29.83\%}. In terms of image quality, IABEdit attains \textbf{37.58\%}, outperforming SuperEdit (\textbf{31.45\%}). The slight variation in performance as compared to the GPT-4o model is observed due to the variation in sample size (i.e., 30 for human evaluation and 560 for GPT-4o evaluation) used for both evaluation methods. Human study results are in the \textcolor{blue}{supplementary}.

\noindent\textcolor{black}{\textbf{Training cost and Limitations:} The limitations and training time comparisons are mentioned in the \textcolor{blue}{supplementary} material.}

\section{Conclusion}

\textcolor{black}{IABEdit reframes instruction-guided editing as a problem of semantic alignment rather than conditioning. The argument is simple, a generative editor should not be trusted to follow an instruction it was never asked to semantically verify. By distilling spatially-aware descriptors from a frozen vision-language model and turning the residual into a gradient, IABEdit makes instruction satisfaction part of the training objective itself. Adherence is no longer assumed, it is measured, and minimized. Because this signal acts only during training, it leaves inference untouched and transfers across architectures, strengthening U-Net and MMDiT backbones alike. It even surpasses a proprietary baseline on the hardest case we tested, identity preservation under heavy occlusion, where curated benchmarks give way to real-world conditions. More broadly, IABEdit shows that gradient-aligned VLM distillation is a general way to supervise generation against meaning, not just pixels. The same principle extends naturally to video editing, 3D generation, and multimodal content creation, wherever verifying what was produced matters as much as producing it.}

\section*{Acknowledgment}
\textcolor{black}{This research was supported by the FAE team, AMD India, through GPU resources and by the IndiaAI Mission and Meta through the Srijan: Centre of Excellence for Generative AI. Chiranjeev is partially supported through the PMRF Fellowship.}

\bibliographystyle{splncs04}
\bibliography{main}

\begin{thebibliography}{10}
\providecommand{\url}[1]{\texttt{#1}}
\providecommand{\urlprefix}{URL }
\providecommand{\doi}[1]{https://doi.org/#1}

\bibitem{bai2025qwen25vltechnicalreport}
Bai, S., Chen, K., Liu, X., Wang, J., Ge, W., Song, S., Dang, K., Wang, P., Wang, S., Tang, J., Zhong, H., Zhu, Y., Yang, M., Li, Z., Wan, J., Wang, P., Ding, W., Fu, Z., Xu, Y., Ye, J., Zhang, X., Xie, T., Cheng, Z., Zhang, H., Yang, Z., Xu, H., Lin, J.: Qwen2.5-vl technical report. arXiv preprint arXiv:2502.13923  (2025)

\bibitem{brooks2023instructpix2pix}
Brooks, T., Holynski, A., Efros, A.A.: Instructpix2pix: Learning to follow image editing instructions. In: CVPR. pp. 18392--18402 (2023)

\bibitem{caron2021emerging}
Caron, M., Touvron, H., Misra, I., J{\'e}gou, H., Mairal, J., Bojanowski, P., Joulin, A.: Emerging properties in self-supervised vision transformers. In: ICCV. pp. 9630--9640 (2021)

\bibitem{comanici2025gemini}
Comanici, G., Bieber, E., Schaekermann, M., Pasupat, I., Sachdeva, N., Dhillon, I., Blistein, M., Ram, O., Zhang, D., Rosen, E., et~al.: Gemini 2.5: Pushing the frontier with advanced reasoning, multimodality, long context, and next generation agentic capabilities. arXiv preprint arXiv:2507.06261  (2025)

\bibitem{couairon2023diffedit}
Couairon, G., Verbeek, J., Schwenk, H., Cord, M.: Diffedit: Diffusion-based semantic image editing with mask guidance. ICLR  (2023)

\bibitem{deng2025emerging}
Deng, C., Zhu, D., Li, K., Gou, C., Li, F., Wang, Z., Zhong, S., Yu, W., Nie, X., Song, Z., Shi, G., Fan, H.: Emerging properties in unified multimodal pretraining. arXiv preprint arXiv:2505.14683  (2025)

\bibitem{deng2019arcface}
Deng, J., Guo, J., Xue, N., Zafeiriou, S.: Arcface: Additive angular margin loss for deep face recognition. In: CVPR. pp. 4685--4694 (2019)

\bibitem{dhariwal2021diffusion}
Dhariwal, P., Nichol, A.: Diffusion models beat gans on image synthesis. NeurIPS  \textbf{34},  8780--8794 (2021)

\bibitem{esser2024scaling}
Esser, P., Kulal, S., Blattmann, A., Entezari, R., M\"{u}ller, J., Saini, H., Levi, Y., Lorenz, D., Sauer, A., Boesel, F., Podell, D., Dockhorn, T., English, Z., Rombach, R.: Scaling rectified flow transformers for high-resolution image synthesis. In: ICML. pp. 12606--12633 (2024)

\bibitem{fu2024guiding}
Fu, T.J., Hu, W., Du, X., Wang, W., Yang, Y., Gan, Z.: Guiding instruction-based image editing via multimodal large language models. In: ICLR. pp. 54820--54833 (2024)

\bibitem{geng2024instructdiffusion}
Geng, Z., Yang, B., Hang, T., Li, C., Gu, S., Zhang, T., Bao, J., Zhang, Z., Li, H., Hu, H., Chen, D., Guo, B.: Instructdiffusion: A generalist modeling interface for vision tasks. In: CVPR. pp. 12709--12720 (2024)

\bibitem{gu2025multi}
Gu, X., Li, M., Zhang, L., Chen, F., Wen, L., Luo, T., Zhu, S.: Multi-reward as condition for instruction-based image editing. In: ICLR. pp. 82243--82262 (2025)

\bibitem{ho2020denoising}
Ho, J., Jain, A., Abbeel, P.: Denoising diffusion probabilistic models. NeurIPS  \textbf{33},  6840--6851 (2020)

\bibitem{hu2022lora}
Hu, E.J., Shen, Y., Wallis, P., Allen-Zhu, Z., Li, Y., Wang, S., Wang, L., Chen, W.: Lo{RA}: Low-rank adaptation of large language models. In: ICLR (2022)

\bibitem{huang2024smartedit}
Huang, Y., Xie, L., Wang, X., Yuan, Z., Cun, X., Ge, Y., Zhou, J., Dong, C., Huang, R., Zhang, R., Shan, Y.: Smartedit: Exploring complex instruction-based image editing with multimodal large language models. In: CVPR. pp. 8362--8371 (2024)

\bibitem{hui2025hqedit}
Hui, M., Yang, S., Zhao, B., Shi, Y., Wang, H., Wang, P., Xie, C., Zhou, Y.: Hq-edit: A high-quality dataset for instruction-based image editing. In: International Conference on Learning Representations. pp. 73407--73424 (2025)

\bibitem{hurst2024gpt}
Hurst, A., Lerer, A., Goucher, A.P., Perelman, A., Ramesh, A., Clark, A., Ostrow, A., Welihinda, A., Hayes, A., Radford, A., et~al.: Gpt-4o system card. arXiv preprint arXiv:2410.21276  (2024)

\bibitem{labs2025flux}
Labs, B.F., Batifol, S., Blattmann, A., Boesel, F., Consul, S., Diagne, C., Dockhorn, T., English, J., English, Z., Esser, P., Kulal, S., Lacey, K., Levi, Y., Li, C., Lorenz, D., Müller, J., Podell, D., Rombach, R., Saini, H., Sauer, A., Smith, L.: Flux. 1 kontext: Flow matching for in-context image generation and editing in latent space. arXiv preprint arXiv:2506.15742  (2025)

\bibitem{li2025superedit}
Li, M., Gu, X., Chen, F., Xing, X., Wen, L., Chen, C., Zhu, S.: Superedit: Rectifying and facilitating supervision for instruction-based image editing. In: ICCV. pp. 19206--19215 (2025)

\bibitem{controlnet_plus_plus}
Li, M., Yang, T., Kuang, H., Wu, J., Wang, Z., Xiao, X., Chen, C.: Controlnet++: Improving conditional controls with efficient consistency feedback. In: ECCV. pp. 129--147 (2024)

\bibitem{liu2023visual}
Liu, H., Li, C., Wu, Q., Lee, Y.J.: Visual instruction tuning. NeurIPS  \textbf{36},  34892--34916 (2023)

\bibitem{manchanda2023d}
Manchanda, S., Bhagwatkar, K., Balutia, K., Agarwal, S., Chaudhary, J., Dosi, M., Chiranjeev, C., Vatsa, M., Singh, R.: D-lord: Dysl-ai database for low-resolution disguised face recognition. T-BIOM  \textbf{6}(2),  147--157 (2024)

\bibitem{nichol2022glide}
Nichol, A.Q., Dhariwal, P., Ramesh, A., Shyam, P., Mishkin, P., Mcgrew, B., Sutskever, I., Chen, M.: Glide: Towards photorealistic image generation and editing with text-guided diffusion models. In: ICML. pp. 16784--16804 (2022)

\bibitem{pan2024kosmos}
Pan, X., Dong, L., Huang, S., Peng, Z., Chen, W., Wei, F.: Kosmos-g: Generating images in context with multimodal large language models. In: ICLR. pp. 40959--40974 (2024)

\bibitem{podell2024sdxl}
Podell, D., English, Z., Lacey, K., Blattmann, A., Dockhorn, T., M{\"u}ller, J., Penna, J., Rombach, R.: Sdxl: Improving latent diffusion models for high-resolution image synthesis. In: ICLR. pp. 1862--1874 (2024)

\bibitem{radford2021learning}
Radford, A., Kim, J.W., Hallacy, C., Ramesh, A., Goh, G., Agarwal, S., Sastry, G., Askell, A., Mishkin, P., Clark, J., Krueger, G., Sutskever, I.: Learning transferable visual models from natural language supervision. In: ICML. pp. 8748--8763 (2021)

\bibitem{ramesh2022hierarchical}
Ramesh, A., Dhariwal, P., Nichol, A., Chu, C., Chen, M.: Hierarchical text-conditional image generation with clip latents. arXiv preprint arXiv:2204.06125  (2022)

\bibitem{ramesh2021zero}
Ramesh, A., Pavlov, M., Goh, G., Gray, S., Voss, C., Radford, A., Chen, M., Sutskever, I.: Zero-shot text-to-image generation. In: ICML. pp. 8821--8831 (2021)

\bibitem{rombach2022high}
Rombach, R., Blattmann, A., Lorenz, D., Esser, P., Ommer, B.: High-resolution image synthesis with latent diffusion models. In: CVPR. pp. 10674--10685 (2022)

\bibitem{saharia2022photorealistic}
Saharia, C., Chan, W., Saxena, S., Li, L., Whang, J., Denton, E.L., Ghasemipour, K., Gontijo~Lopes, R., Karagol~Ayan, B., Salimans, T., Ho, J., Fleet, D.J., Norouzi, M.: Photorealistic text-to-image diffusion models with deep language understanding. NeurIPS  \textbf{35},  36479--36494 (2022)

\bibitem{sheynin2024emu}
Sheynin, S., Polyak, A., Singer, U., Kirstain, Y., Zohar, A., Ashual, O., Parikh, D., Taigman, Y.: Emu edit: Precise image editing via recognition and generation tasks. In: CVPR. pp. 8871--8879 (2024)

\bibitem{singh2024smartmask}
Singh, J., Zhang, J., Liu, Q., Smith, C., Lin, Z., Zheng, L.: Smartmask: Context aware high-fidelity mask generation for fine-grained object insertion and layout control. In: CVPR. pp. 6497--6506 (2024)

\bibitem{xie2023smartbrush}
Xie, S., Zhang, Z., Lin, Z., Hinz, T., Zhang, K.: Smartbrush: Text and shape guided object inpainting with diffusion model. In: CVPR. pp. 22428--22437 (2023)

\bibitem{zhang2023magicbrush}
Zhang, K., Mo, L., Chen, W., Sun, H., Su, Y.: Magicbrush: A manually annotated dataset for instruction-guided image editing. NeurIPS  \textbf{36},  31428--31449 (2023)

\bibitem{zhang2024hive}
Zhang, S., Yang, X., Feng, Y., Qin, C., Chen, C.C., Yu, N., Chen, Z., Wang, H., Savarese, S., Ermon, S., Xiong, C., Xu, R.: Hive: Harnessing human feedback for instructional visual editing. In: CVPR. pp. 9026--9036 (2024)

\bibitem{zhao2024ultraedit}
Zhao, H., Ma, X., Chen, L., Si, S., Wu, R., An, K., Yu, P., Zhang, M., Li, Q., Chang, B.: Ultraedit: Instruction-based fine-grained image editing at scale. NeurIPS  \textbf{37},  3058--3093 (2024)

\end{thebibliography}

%

\clearpage
\begin{center}
{\Large\bfseries Supplementary Material}
\end{center}

The supplementary is structured as follows:
\begin{itemize}
    \item \textbf{Section \ref{sec:flux_maths}: \textit{IABEdit with FLUX}.} \emph{(In reference to Section 2.1 of the main paper)} \\
    
    \item \textbf{Section \ref{sec:human_study_qualitative}: \textit{Qualitative Results of Human study.}} \emph{(In reference to Section 4 (Human Study) of the main paper)} \\

    \item \textbf{Section \ref{sec:facial_ablation_surv}:\textit{ Ablation Results on Facial Disguise Image Editing and Surveillance Scenarios.}} \emph{(In reference to Section 4 (Real-world Surveillance Scenarios) of the main paper)} \\

    \item \textbf{Section \ref{sec:implement_details}: \textit{Implementation Details.}} \emph{(In reference to Section 4, Figure 3 of the main paper)} \\

    $\bullet$ Section \ref{sec:sdv15_protocol}: Diffusion Model: Stable Diffusion v1.5 \\

    $\bullet$ Section \ref{sec:flux_protocol}: Flow Matching Model: FLUX.1-dev \\

    $\bullet$ Section \ref{sec:instruction_prompt}: Editing Instruction-based Prompts \\

    \item \textbf{Section \ref{sec:qualitative_results}: \textit{Qualitative Results.}} \emph{(In reference to Section 3 (Architecture and Implementation Details) of the main paper)} \\

    $\bullet$ Section \ref{sec:gradient_maps}: VLM-focussed Gradient Maps Visualization \\

    $\bullet$ Section \ref{sec:magic_brush_qualitative}: Qualitative Results on MagicBrush Benchmark \\

    $\bullet$ Section \ref{sec:complex_instruction_results}: Results on Complex Instruction Scenarios \\

    \item \textbf{Section \ref{sec:quant_rebuttal_results}: \textit{Additional Quantitative Results.}} \emph{(In reference to Section 4 of the main paper)} \\

    $\bullet$ Section \ref{sec:masked_cs_p}: Masked CS-P evaluation on RealEdit dataset \\

    $\bullet$ Section \ref{sec:emuedit_test_set_results}: Results on EMU Edit test set \\

    \item \textbf{Section \ref{sec:image_quality_comparison}: \textit{Discussion on Image Quality.}} \emph{(In reference to Section 4 (Agent-Based and Human-Centric Evaluation) of the main paper)} \\

    \item \textbf{Section \ref{sec:training_cost}: \textit{Training Cost.}} \emph{(In reference to Section 4 (Training cost and Limitations) of the main paper)} \\

    \item \textbf{Section \ref{sec:limitations}: \textit{Limitations and Failure modes.}} \emph{(In reference to Section 4 (Training cost and Limitations) of the main paper)} \\
   
\end{itemize}

\section{IABEdit with FLUX}
\label{sec:flux_maths}

In the main paper, the proposed IABEdit framework is shown with noise-prediction diffusion models. However, our IABEdit framework is model-agnostic and is also compatible with flow-based generative models such as FLUX. In the FLUX.1-dev model, generation is formulated as a continuous flow between Gaussian noise and the clean data distribution, where the model learns a velocity field that progressively transforms a noisy latent toward the clean latent representation.

Given a clean latent representation $z_0$ and Gaussian noise $\epsilon \sim \mathcal{N}(0,I)$, the forward interpolation process constructs an intermediate latent $z_t$ as:

\begin{equation}
z_t = (1 - t) z_0 + t \epsilon, \quad \epsilon \sim \mathcal{N}(0, I)
\end{equation}

In flow-matching models, it is implemented using a multi-modal diffusion transformer (MMDiT) in FLUX,which predicts a velocity field $v_\theta$ that transforms the latent state toward the clean data manifold. In IABEdit for FLUX model, the conditioning consists of the textual instruction embedding from the CLIP-ViT-L/14 text encoder
($c_{\text{ins}}$), T5-XXL text encoder ($c_{{t5}_{\text{ins}}}$), the source-image embedding ($c_{\text{src}}$), and the source latent ($s$) used for context and structural preservation:

\begin{equation}
c_{\text{ins}} = \text{CLIP}(\text{instruction}); \,
c_{{t5}_{\text{ins}}} = \text{T5}(\text{instruction}); \,
c_{\text{src}} = \text{MLP}(\phi(I_{\text{src}}))
\end{equation}

The velocity predictor is trained using the standard flow-matching objective $\mathcal{L}_{\text{N}}$:

\begin{equation}
\mathcal{L}_{\text{N}} = \mathbb{E}_{t,s,c} \left[ \left\|v_\theta(z_t \oplus s, c, t) - (\epsilon- z_0) \right\|_2^2\right]
\end{equation}

where $c = (c_{\text{ins}}, c_{\text{src}}, c_{{t5}_{\text{ins}}})$ denotes the combined conditioning signal, $z_t$ denotes the noisy latent at timestep $t$, and $\oplus$ represents latent concatenation with the source context. During edited image latent generation, the predicted velocity provides a one-step estimate of the clean latent by integrating the flow field:

\begin{equation}
z_0 \approx z_0' = z_t - t \, v_\theta((z_t \oplus s), c_{\text{ins}}, c_{\text{src}}, c_{{t5}_{\text{ins}}}, t).
\end{equation}

This one-step estimate is employed for small timesteps $t \leq \tau$ ($\tau$=200), to obtain an approximate ($z_0'$) of clean latent ($z_0$). This one-step latent $z'_{0}$ is passed to the VAE decoder ($\mathcal{D}_{\text{VAE}}$) to obtain the generated edited image ($I_{G_{edit}} = \mathcal{D}_{\text{VAE}}(z'_{0})$), which is passed as an input to the VLM (LLaVA) model to analyze the generated edited image and provide gradient-based feedback to the FLUX-model for performing precise localised edits.

Therefore, this formulation enables IABEdit to operate seamlessly with flow-based backbones such as FLUX, while retaining the structured conditioning and gradient-based semantic alignment mechanisms introduced in our framework. 


\section{Human Study Qualitative Results}
\label{sec:human_study_qualitative}

\begin{figure*}[h]
    \centering
    \includegraphics[width=0.98\textwidth]{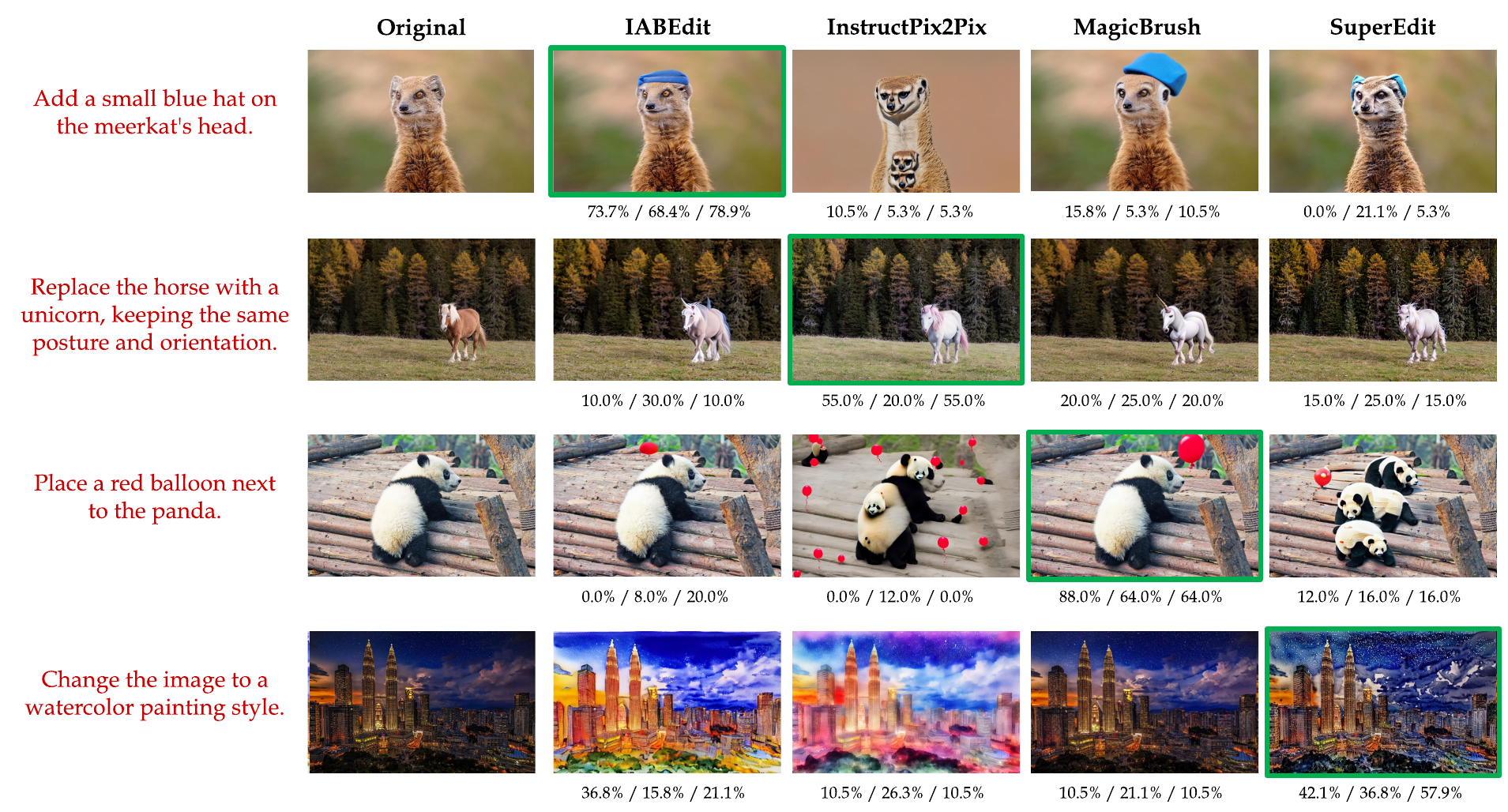}
    \caption{Illustrating the overall \textcolor{darkgreen}{best performing models} according to the user responses for a few sets of samples. Below every image, it depicts the percentage of people saying it is the best-performing model in terms of Instruction Following/Non-editing required region Preservation/Quality of Generated image.}
    \label{fig:quality_realedit_response}
\end{figure*}

Figure \ref{fig:quality_realedit_response} illustrates qualitative examples alongside corresponding user study responses for a representative set of samples. It showcases the best-performing model for each sample, with the percentage of people saying it is the best-performing model in terms of Instruction Following, Preserving, and Quality. 

\textit{Human Study Protocol:} We conducted a human study on the RealEdit dataset with 30 participants aged 20-45 to evaluate image editing performance from a human-centric perspective. Volunteer participants who are not involved in the proposed research were chosen. Each participant evaluated 30 randomly sampled instances, each consisting of the original image, editing instruction, and edited outputs from different models. The evaluation was performed in a blinded setting, where model identities were hidden and outputs were presented in random order. Participants selected the best-performing model based on instruction following, preservation of non-instructed regions, and overall image quality. Finally, the results were obtained by aggregating participant values across all 30 samples and the complete per-sample breakdown.


\section{Facial Image Editing and its impact on Surveillance scenarios}
\label{sec:facial_ablation_surv}
This section discusses the significance of each component used for performing image editing tasks in facial disguise settings and the effectiveness of facial editing in enhancing surveillance scenarios. 

\subsection {Ablation Results on Facial Disguise Image Editing}

Table \ref{tab:ablation_dlord} highlights the significance of each component of the IABEdit method by assessing based on the ablation results on the facial disguise editing task on the real-world D-LORD face dataset. We achieve the higher HM scores as the model is trained by introducing each new loss component. The results in Table \ref{tab:ablation_dlord} show that training the IABEdit framework with all Denoising ($\mathcal{L}_{N}$), Distillation ($\mathcal{L}_{D}$) and coherence ($\mathcal{L}_{C}$) loss functions, together leads to the best balanced image editing with achieving the highest HM score of 46.74  and DINO-preserving score of 55.33. The combined inclusion of all components significantly improves the model's editing capability (CS-P: +11.23, CLIP-T: +2.43, HM: +5.03 and DINO-P: +4.92) as compared to the model trained with only Diffusion loss ($\mathcal{L}_{N}$). As each new loss component is introduced for training the IABEdit framework, its performance improves consistently. Therefore, this signifies the importance of the distillation component and the facial identity component, which are enhanced by gradient-enabled meaningful supervisory signals. This strengthens the diffusion model’s ability to produce accurate, semantically aligned edits from natural language instructions. 

\begin{table}[t!]\renewcommand{\arraystretch}{1.2}
\centering
\caption{\label{tab:ablation_dlord}Ablation study on D-LORD dataset for disguised facial image editing task.} 
\resizebox{0.7\linewidth}{!}{
\begin{tabular}{l|c|c|c|c}
\hline
\textbf{Loss Component} & \textbf{CS-P} $\uparrow$ & \textbf{CLIP-T} $\uparrow$ & \textbf{HM} $\uparrow$ & \textbf{DINO-P} $\uparrow$ \\ \hline
$\boldsymbol{\mathcal{L}_{N}}$ & 68.70 & 30.67 & 41.71 & 50.41 \\
 \textbf{$\mathcal{L}_{N} + \mathcal{L}_{D}$} & 74.64 & 31.95 & 44.68 & 53.59\\ 
$\boldsymbol{\mathcal{L}_{N} + \mathcal{L}_{D} + \mathcal{L}_{C}}$ & \textbf{79.93} & \textbf{33.10} & \textbf{46.74} & \textbf{55.33}\\ 
\hline
\end{tabular}}
\end{table}

\subsection{Surveillance Results in Low-Resolution Setting}

\begin{figure*}[t!]
    \centering
    \includegraphics[width=0.88\textwidth]{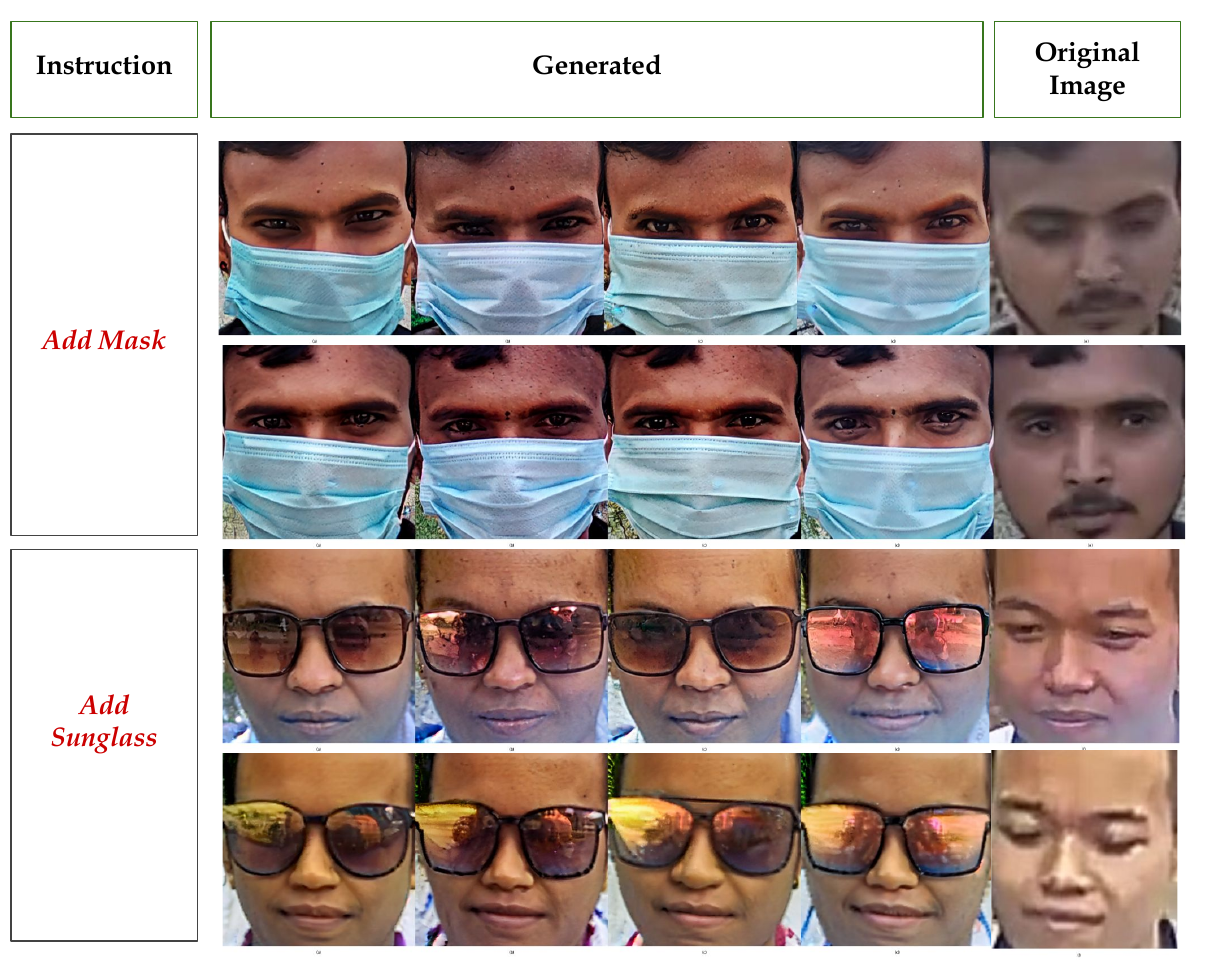}
    \caption{Visualizing the generated edited samples for low-resolution (LR) real-world surveillance face images on D-LORD dataset. The original LR images are edited based on the instructions: (i) "Add a face mask to the person’s face, which covers the nose and mouth region, obscuring the lower half of the face." and (ii) "Add sunglasses to the person’s face, which covers the eyes and upper part of the nose region.".
 }
    \label{fig:quality_LR_dlord}
\end{figure*}

\begin{figure}[t!]
    \centering
    \includegraphics[width=0.7\linewidth]{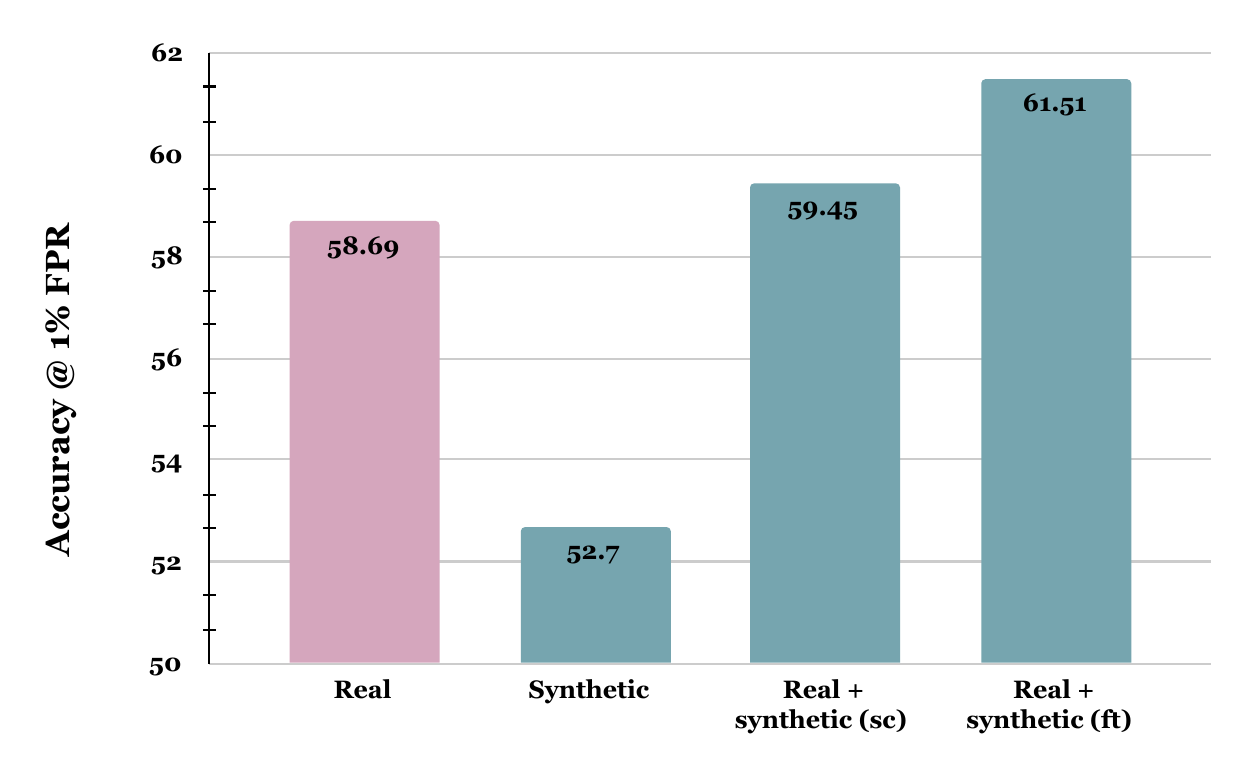}
    \caption{Recognition performance across training setups using real and IABEdit-generated face pairs. Fine-tuning on combined data yields the best accuracy, showing the value of instruction-based editing for occluded face recognition.}
    \label{fig:accuracy}
\end{figure}

Figure \ref{fig:quality_LR_dlord} illustrates that generated edited facial outputs accurately adhere to the given instructions, achieving precise localization of occlusions on the target facial regions. The facial mask editing examples exhibit strong coverage of the lower half of the face without significant distortion of surrounding regions, indicating effective structural consistency. Similarly, the sunglasses-editing examples maintain correct alignment with the eye region while preserving facial attributes such as structure and texture across the non-edited regions, reflecting high identity retention.

In terms of visual quality, the generated edits demonstrate natural blending with minimal artifacts, despite the low-resolution nature of the original surveillance images. The ability to preserve the inherent characteristics of the unedited regions while introducing realistic occlusions highlights the robustness of the proposed IABEdit framework. It effectively handles real-world face editing scenarios even in low-resolution settings by maintaining strong instruction adherence and high visual quality in the generated results.

These synthetically edited images are then used to train a face recognition model, highlighting how semantically controlled editing can enhance biometric systems under occlusion. We generate a training dataset comprising 40,000 positive image pairs and 120,000 negative pairs each for real and synthetic data. A ResNet-18 model is trained using a contrastive loss objective, with two settings: training from scratch (sc) and fine-tuning (ft) on the edited data. Our test set includes images at varied distances and having variations in occlusions covering different regions of face captured from surveillance camera, showcasing real-world surveillance scenarios.

As shown from the results mentioned in Figure \ref{fig:accuracy} , the model trained on only real data achieves 58.69\% accuracy @1\% FPR. Incorporating only synthetic data yields 52.7\%, reflecting the challenge of using synthetic images alone. However, combining both real and synthetic data and training from scratch improves performance to 59.45\%, suggesting that synthetic data adds complementary variance. Notably, fine-tuning the recognition model on the combined dataset leads to the best performance at 61.51\%, showcasing that IABEdit-generated edits help improve recognition robustness by simulating realistic facial variations.

These results confirm the practical applicability of IABEdit in surveillance-driven biometric pipelines. The model enables high-quality, instruction-guided face edits that can simulate challenging real-world occlusions. When used for training, these edits enhance recognition accuracy, offering a powerful tool for augmenting face datasets in low-visibility and occlusion-heavy environments.

\section{Implementation Details}
\label{sec:implement_details}
In this section, we provide the implementation details of the components utilized in the proposed algorithm. We also make the resources available for the research community at this link: \url{https://github.com/IAB-IITJ/IABEdit}

\subsection{Diffusion Model: Stable Diffusion v1.5}
\label{sec:sdv15_protocol}

IABEdit is trained in a latent diffusion technique (LDM) for generation purposes, and the LLaVA-7B (LLaVA-v1.6-Mistral-7B) model is used as a Descriptive Anchor and Instruction Aligner for the semantic-supervised distillation process. Stable Diffusion v1.5 weights are used as the pre-trained weights for the editing diffusion model with the UNet's denoising architecture. LDM is conditioned with the textual instruction obtained from CLIP text-encoder, and image conditioning is done with CLIP (ViT-B/32) image-encoder in case of scene editing and face recognition (ResNet-18) model pre-trained with ArcFace loss for face editing. Also, we enable classifier-free diffusion guidance for both image and text conditions. The diffusion model operates on an input image of size $512 \times 512$, whereas LLaVA processes the generated denoised edited image after resizing it to $336 \times 336$ dimensions. The hyperparameter setting for training the IABEdit framework includes the learning rate of 1e-5, weight decay of 1e-2, a training batch size of 2, and a \emph{constant with warmup} scheduler is utilized with a warmup ratio of 500 steps. \textcolor{black}{The temperature $T$ in $\mathcal{L}_{\text{D}}$ loss in Equation 3 (in the main paper) is set to 5.0.} Overall, the IABEdit is trained for 118400 training steps for the scene editing task and 58000 training steps for the face editing task. While $\lambda_{N}$, $\lambda_{D}$ and $\lambda_{C}$ are kept as 1.0. The timestep scheduler threshold $\tau$ is set to 200, as the latent samples with timesteps less than 200 are less noisy. The predicted original latent sample from these noisy latents, using one-step denoising, is highly similar to the ground-truth original latent. The inference stage remains VLM free. The DDIM noise scheduler is utilized with the inference sampling steps of 30 and has a text and image guidance of 7.5 and 2.0, respectively. All comparing instruction-based editing models are evaluated at their peak performances with their best setting combination of sampling steps, text and image guidance. Further, for performing ablation with another strong VLM, we have utilized Qwen2.5-VL (Qwen2.5-VL-7B-Instruct).

\begin{table}[t!]\renewcommand{\arraystretch}{1.0}
\centering
\caption{\label{tab:tau_threshold} Demonstrating results based on single-step diffusion-based denoising at different timestep thresholds $\tau$.}
\resizebox{0.45\textwidth}{!}{
\begin{tabular}{l|c|c|c}
\hline
$\tau$ & \textbf{CS-P} $\uparrow$ & \textbf{CLIP-T} $\uparrow$ & \textbf{HM} $\uparrow$ \\ \hline
\textbf{200} & 89.65 & 27.60 & 42.00 \\
\textbf{300} & 93.77 & 26.13 & 40.67 \\ 
\hline
\end{tabular}}
\end{table}

Table \ref{tab:tau_threshold} presents the image editing performance on the RealEdit benchmark when selecting different timestep threshold ($\tau$). These results reflect the selection of ($\tau$) to perform efficient latent estimation ($z'_{0}$) via a single-step denoising step as shown in Equation $2$ in Section $2.2$. $\tau=200$ yields the highest editing performance (CLIP-T and HM) , demonstrating better instruction following with balanced editing (HM=42.00) compared to performance values at $\tau=300$ reflects under-editing (less instruction adherence. Increasing $\tau$ to 300 results in higher CS-P (93.77) because fewer image regions are modified, leading to stronger content preservation but weaker instruction adherence (CLIP-T: 26.13). Consequently, the overall harmonic mean (HM) decreases, indicating under-editing. Therefore, $\tau=200$ provides the best balance between faithful instruction following and context preservation. The latent at $\tau=200$ is less noisy, hence iterative denoising can be replaced with denoising in a single step, and further the resulting latent can be converted to an image using the VAE decoder. The final single-step denoised image is used as input to other models, such as VLMs. These results led to the choice of selecting $\tau=200$ as the latent $z_{t}$ (t $\leq$ $\tau$) can be used to compute the latent $z'_{0}$ using a simple formulation (mentioned in Equation $2$ in the main paper), which is very similar to the original latent $z_{0}$.


\begin{figure}[t!]
    \centering
    \includegraphics[width=0.7\linewidth]{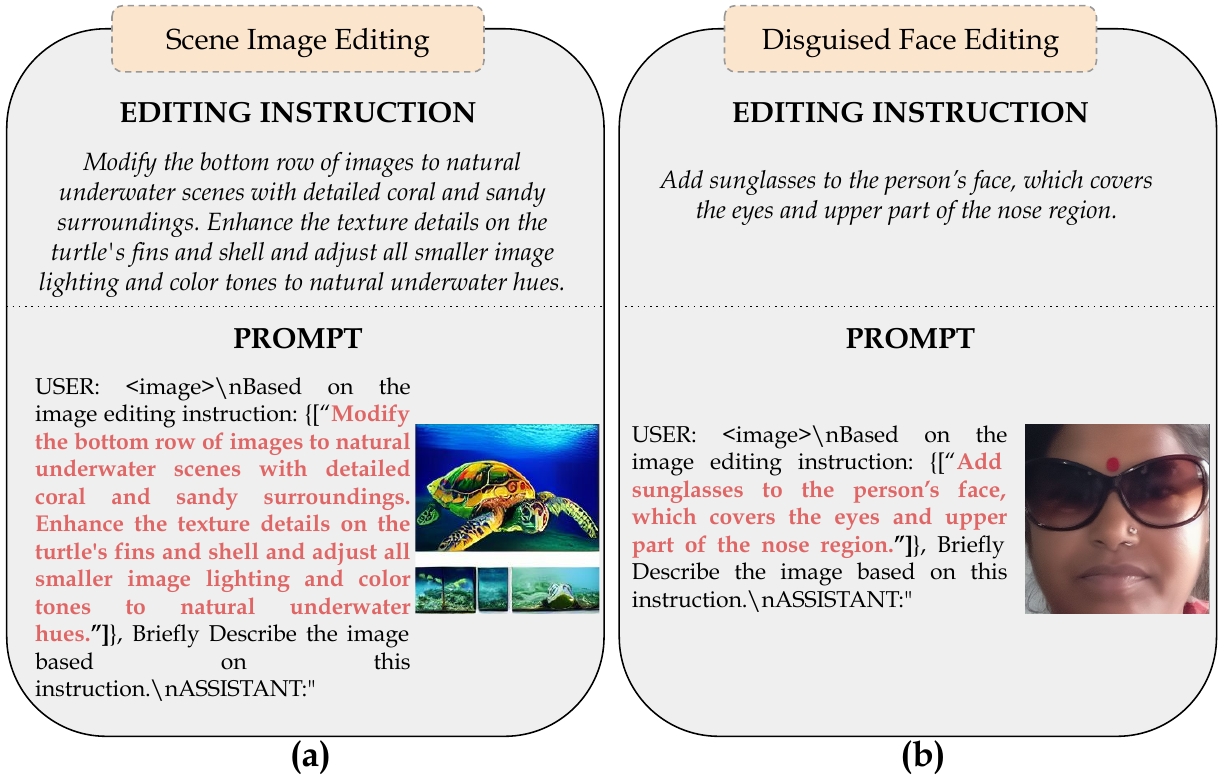}
    \caption{Examples of editing instruction-following based prompts for vision language models (VLMs), on (a) scenes and (b) facial datasets.}
    \label{fig:vlm_prompt}
\end{figure}

\subsection{Flow Matching Model: FLUX.1}
\label{sec:flux_protocol}

FLUX-based image editing is built on a FLUX.1-dev base model, which adopts the Multi-Modal Diffusion Transformer (MMDiT) architecture. The proposed IABEdit + FLUX.1 model is trained using backward gradient supervision derived from the VLM LLaVA, enabling instruction-aligned editing through VLM-guided gradient flow. The model is trained for 26,000 iterations on the SuperEdit-40K dataset. This dataset consists of paired images obtained from several existing editing datasets introduced in prior works, including InstructPix2Pix, MagicBrush, and the Seed-Data-Edit Dataset, which provide diverse instruction-guided image editing samples. This FLUX-based image editing is performed for the scene image editing task trained with denoising and distillation loss for $\tau$ = 200. 

\noindent The IABEdit method with the \textit{Diffusion and VLMs-based} architecture is trained on an 80 GB A100 GPU, while the computationally heavy \textit{FLUX and VLMs-based} architecture is trained on MI325X AMD GPUs.

\subsection{Editing Instruction-based Prompts}
\label{sec:instruction_prompt}

Figure \ref{fig:vlm_prompt} presents sample VLM prompt inputs for both scene and facial editing scenarios. These prompts are constructed directly from the underlying editing instructions. This design allows the VLMs to better interpret the semantic intent behind each editing instruction. It enables them to explicitly describe how precisely the diffusion model performed the spatially localized edits by adhering to the requested intended modifications while preserving the regions that were not meant to be edited.

\section{Additional Qualitative Results}
\label{sec:qualitative_results}

\subsection{VLM-focussed Gradient Maps Visualization}
\label{sec:gradient_maps}

\begin{figure*}[t!]
    \centering
    \includegraphics[width=0.98\textwidth]{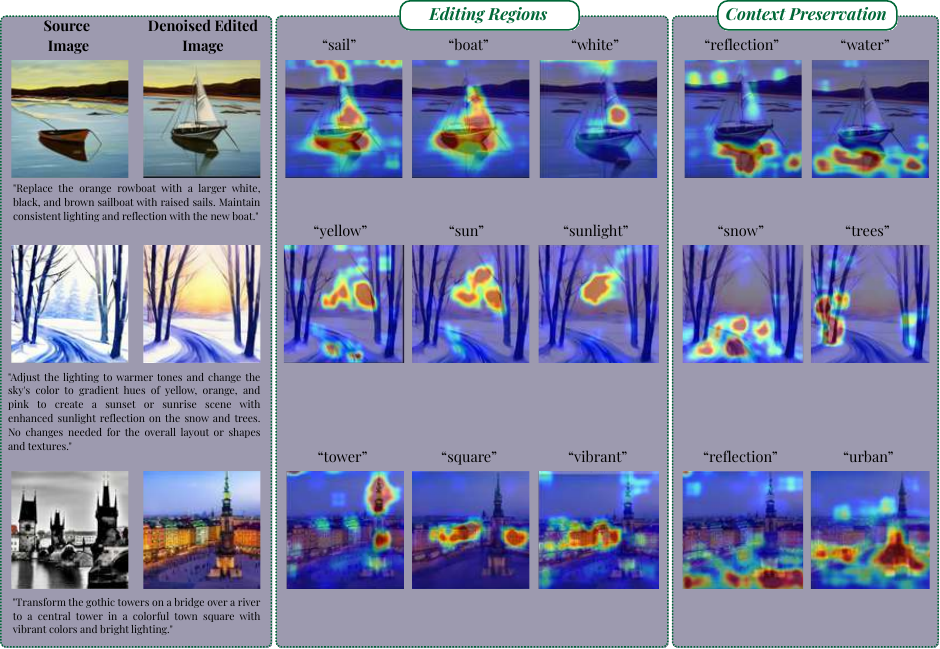}
    \caption{Grad-CAM visualization for LLaVA based gradient backprop showcasing token-level attention during instruction-guided image editing. The figure highlights editing regions (e.g., sail, boat, tower, square) where the model focuses to perform semantic modifications, and context preservation regions (e.g., reflection, water, trees, urban) where attention remains on the surrounding environment to maintain scene structure and visual consistency during editing.}
    \label{fig:gradcam_llava}
\end{figure*}

Illustrating the backward gradient flow from VLM (LLaVA), which impacts the image regions of the denoised edited image to perform editing for \emph{what} and \emph{where} to edit context, we have shown Grad-CAM visualisation in Figure \ref{fig:gradcam_llava}. These maps highlight how the model localizes semantic concepts from the editing instruction and how it preserves surrounding context during editing. The analysis is interpreted across three main aspects: localization, context preservation, and localized editing behavior.

\begin{enumerate}
    \item \textit{\textbf{Localization of Editing Regions:}} The Grad-CAM maps reflecting editing regions demonstrate that the model precisely localizes regions corresponding to the instructions that drive the editing. 
    \begin{itemize}
        \item In Figure \ref{fig:gradcam_llava} (first row), the boat editing example, attention for tokens such as "sail" and "boat" concentrates on the central object, particularly the hull and sail region where the rowboat is transformed into a sailboat. Similarly, the "white" instruction token produces a localized response over the modified boat surface, indicating that the model associates color modification with the correct spatial region.
        
        \item In Figure \ref{fig:gradcam_llava} (second row), the winter landscape example, tokens like "yellow", "sun", and "sunlight" show strong activation near the sky and illumination region, precisely where the lighting conditions are altered.
        
        \item In Figure \ref{fig:gradcam_llava} (third row), the tower and town square transformation example, tokens such as "tower", "square", and "vibrant" produce attention around architectural structures and the central plaza, confirming that structural changes are localized to the intended area. The square and town keywords make the semantic contextual understanding to edit the bridge scenario with the town square.
    \end{itemize}

    These observations indicate that the model learns semantic token-to-region correspondence, enabling spatially accurate alignment of textual instructions to precise locations in the image.

    \item \textit{\textbf{Context Preservation:}} The "Context Preservation" block in Figure \ref{fig:gradcam_llava} illustrates how the model maintains structural and environmental consistency while performing the required edits.
    \begin{itemize}
        \item The descriptive tokens, such as "reflection" and "water", generated by the VLM model highlight attention to preserve the non-editable regions such as water surfaces in the boat scene, ensuring that reflections remain coherent with the newly inserted sailboat.
        
        \item In the snowy forest example, attention for "snow" and "trees" spreads across the ground and tree structures, indicating that these elements are preserved rather than modified.

        \item In the urban square transformation, context-understanding VLMs' descriptive tokens, such as "reflection" and "urban", attend to buildings and ground structures, maintaining scene layout and environmental realism.
    \end{itemize}

    This demonstrates that the IABEdit framework not only focuses on editing regions but also actively supports contextual elements, helping maintain global scene coherence and physical consistency by leveraging ground-truth edited-image understanding through Descriptive Anchor's distillation and passing this distilled long-contextual descriptive knowledge to the image generative model and the trainable context preserving image conditioning module shown in Figure 2 of the main paper.

    \item \textit{\textbf{Evidence of Localized Editing Behavior:}} The combined attention patterns confirm that the editing process is spatially selective rather than globally destructive with the careful key observations from Figure \ref{fig:gradcam_llava} as follows:
    \begin{itemize}
        \item High attention is concentrated in semantically relevant regions, where modifications occur.
        \item Surrounding areas receive lower but structured attention, ensuring they remain unchanged or minimally affected.

        \item Distinct token activations correspond to different semantic components, indicating compositional editing capability.
    \end{itemize}
    
\end{enumerate}

Overall, Figure \ref{fig:gradcam_llava} demonstrates that the proposed IABEdit framework achieves localized semantic manipulation through VLM-based backward semantic gradient flow along $I_{\text{Gedit}} \rightarrow \mathcal{D}_{\text{VAE}} \rightarrow z'_0 \rightarrow \epsilon_\theta$, together with the proposed context-preserving trainable module. As a result, the denoising network and the context-preserving module become semantically aligned with both the textual instructions and the intended image editing. Therefore, during inference, the image editing model, without relying on a VLM, modifies only the required editing regions while preserving overall scene context consistency.

\subsection{Qualitative Results on MagicBrush Benchmark}
\label{sec:magic_brush_qualitative}

\begin{figure*}[t!]
    \centering
    \includegraphics[width=0.8\textwidth]{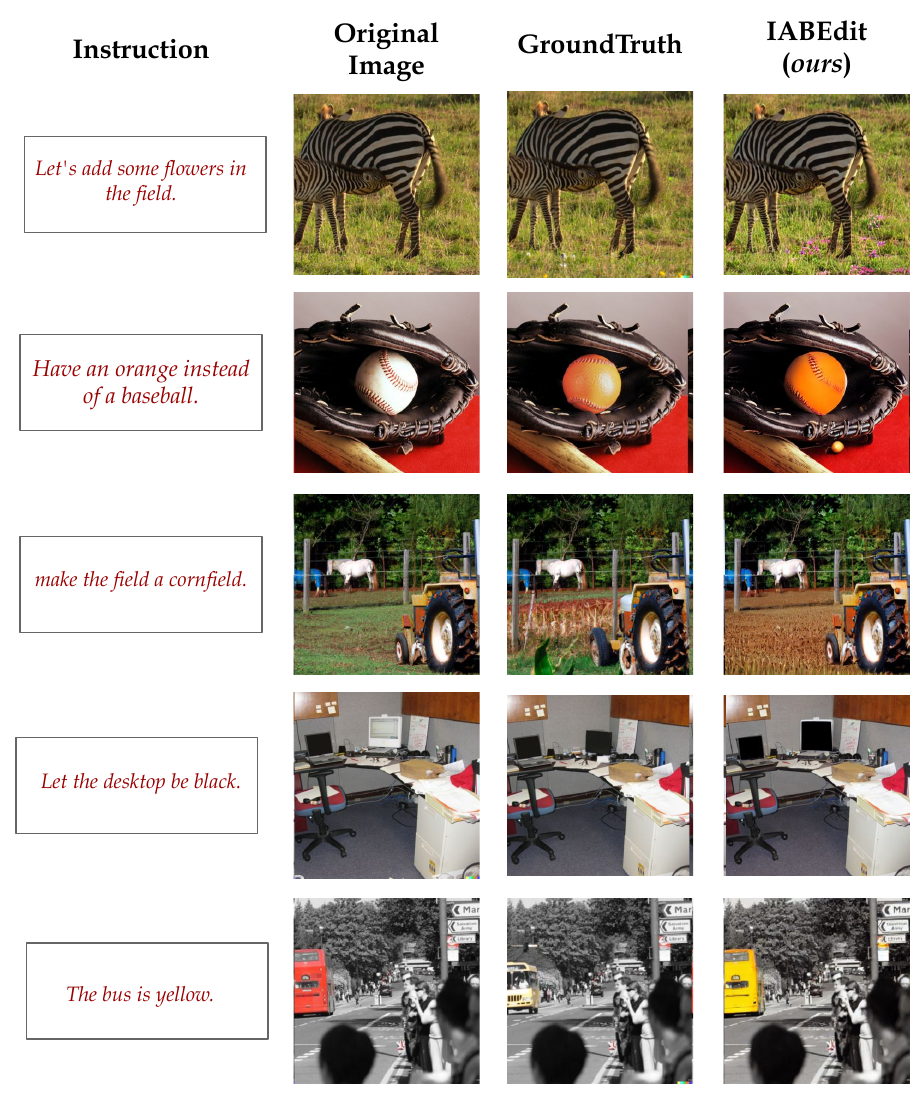}
    \caption{Qualitative comparison of IABEdit on the Magic Brush Dataset. Each row illustrates the comparison between (b) the ground-truth edited image and (c) the image generated by our framework, IABEdit, on (a) the original image. The modifications are done based on the editing instructions shown on the left.}
    \label{fig:quality_magicbrush}
\end{figure*}

Figure \ref{fig:quality_magicbrush} illustrates the qualitative comparisons of IABEdit with respect to the ground truth edits under various editing instructions. Each row presents an original image, the corresponding ground-truth edit, and the output generated by IABEdit. Our framework consistently produces visually coherent and semantically accurate edits aligned with the given instructions.

For instance, in the first example, the instruction "Let’s add some flowers in the field" is correctly interpreted, with IABEdit adding flowers while preserving the zebra and overall scene composition. However, the ground truth image indeed shows only a few small flowers added in the background, whereas IABEdit seems to add flowers more prominently and naturally in the scene. While IABEdit also preserves the fine-grained detail in the image, like the back toe of the zebra. For structural edits such as "make the field a cornfield", IABEdit introduces the new background while preserving foreground objects like the tractor and horse. In the instruction "Let the desktop be black", our method applies localized, very fine changes to the desktop region without distorting surrounding objects. Finally, for "The bus is yellow", the proposed IABEdit model accurately and sharply recolors the bus while preserving other elements of the scene, whereas in the ground truth image, instead of just changing the color of the actual bus, the whole bus is modified. In the second row for instruction: "Have an orange instead of a baseball", our model is not able to replace the baseball with an orange, but it tried to reduce the seam of the ball to make it look like an orange, in contrast, the ground truth image changes the color of the ball to an orange color ball, by creating seams on the original image.

Overall, these results demonstrate that IABEdit is capable of performing localized edits while preserving non-editing context, ensuring that both style and structure remain consistent with the original image.

\subsection{Results on Complex Instruction Scenarios}
\label{sec:complex_instruction_results}

\begin{table*}[t!]
\centering
\caption{\label{tab:complex_instructions}Performance on complex instruction scenarios for RealEdit benchmark dataset. We generated the complex editing instructions for each of the different semantic types of instructions using Gemini-AI Pro 2.5.}
\resizebox{0.9\linewidth}{!}{
\begin{tabular}{l|c|c|c}
\hline
\textbf{Semantic Types in Complex Instructions} & \textbf{CS-P} $\uparrow$ & \textbf{CLIP-T} $\uparrow$ & \textbf{HM} $\uparrow$ \\ 
\hline
\begin{tabular}[c]{@{}l@{}}Direct Simple Instructions (Reward-InstructPix2Pix)\end{tabular} & 89.69 & 27.31 & 41.67 \\
\begin{tabular}[c]{@{}l@{}}Morphological Instructions (Reward-InstructPix2Pix)\end{tabular} & 90.72 & 27.01 & 41.44 \\
\begin{tabular}[c]{@{}l@{}}Lexical Instructions (Reward-InstructPix2Pix)\end{tabular} & 90.43 & 27.43 & 41.93 \\
\begin{tabular}[c]{@{}l@{}}Syntactic Instructions (Reward-InstructPix2Pix)\end{tabular} & 89.55 & 27.67 & 42.09 \\ 
\hline
\begin{tabular}[c]{@{}l@{}}Direct Simple Instructions (SuperEdit)\end{tabular} & 90.93 & 26.01 & 40.24 \\
\begin{tabular}[c]{@{}l@{}}Morphological Instructions (SuperEdit)\end{tabular} & 92.38 & 25.99 & 40.38 \\
\begin{tabular}[c]{@{}l@{}}Lexical Instructions (SuperEdit)\end{tabular} & 92.05 &  26.03 & 40.41 \\
\begin{tabular}[c]{@{}l@{}}Syntactic Instructions (SuperEdit)\end{tabular} & 92.43 & 26.20 & 40.64 \\ 
\hline
\rowcolor{lightred}
\begin{tabular}[c]{@{}l@{}}\textit{Direct Simple Instructions} \textit{(IABEdit)}\end{tabular} & 89.65 & 27.60 & 42.00 \\
\rowcolor{lightred}
\begin{tabular}[c]{@{}l@{}}\textit{Morphological Instructions} \textit{(IABEdit)}\end{tabular} & 89.03 & 27.51 & 41.74 \\
\rowcolor{lightred}
\begin{tabular}[c]{@{}l@{}}\textit{Lexical Instructions} \textit{(IABEdit)}\end{tabular} & 87.83 & 27.69 & 41.79 \\
\rowcolor{lightred}
\begin{tabular}[c]{@{}l@{}}\textit{Syntactic Instructions} \textit{(IABEdit)}\end{tabular} & 87.11 & 27.98 & 42.12 \\ 
\hline
\end{tabular}}
\end{table*} 

To assess the vulnerability of editing models, we evaluated the editing methods across distinct types of complex editing instruction scenarios. We modified the editing instructions based on various semantic types (Direct Simple, Morphological, Lexical and Syntactic instructions), while maintaining the actual meaning of the instructions. We created the modified editing instruction sets benchmark for the complete RealEdit dataset. The modified instructions based on different semantic types are created using the GeminiAI Pro 2.5 model. The \emph{Morphological} types of instructions include words like "non-yellow" and "unfluffy". \emph{Lexical} editing instructions include "empty" and "devoid", while \emph{Syntactic} editing instructions consist of words such as "without" and "do not". Furthermore, the complex instruction scenario-based image editing will be incorporated into the training mechanism by handling them specifically for a wide variety of acceptable prompt-based image editing tasks in future research directions.

\begin{figure*}[t!]
    \centering
    \includegraphics[width=\textwidth]{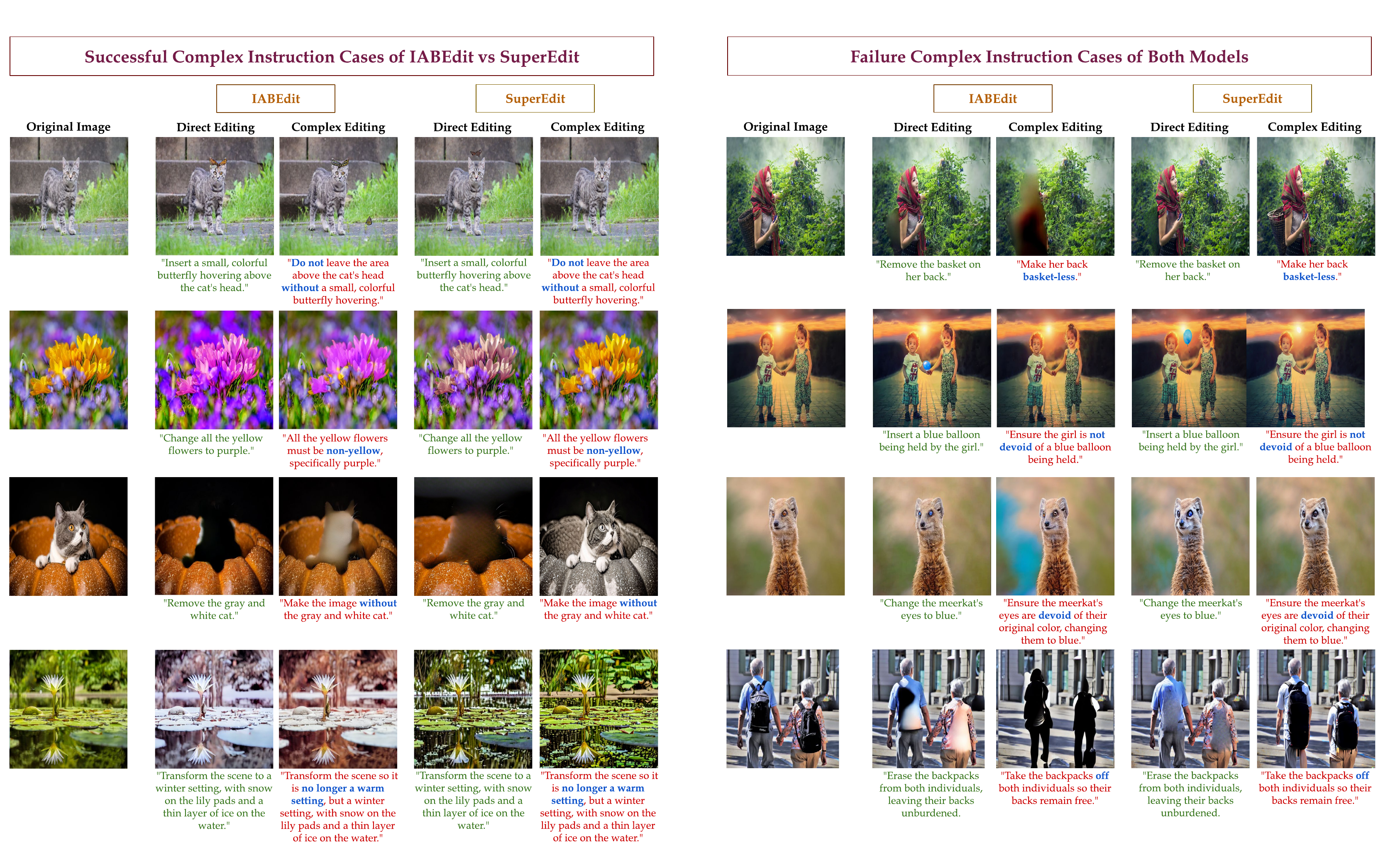}
    \caption{The qualitative results showcasing the \emph{(left part)} successful cases of the IABEdit against SuperEdit in the complex instruction scenarios. The \emph{right part} shows the failure cases where both the models fail with complex editing instructions, where they perform correct edits with the direct simple editing instructions.}
    \label{fig:complex_editing_qulaitative}
\end{figure*}

Table \ref{tab:complex_instructions} shows that different editing methods perform better with direct simple instructions, while IABEdit being the best method (HM=42.00) for balanced editing. Whereas in cases of other types of editing instructions with different semantic types but the same meaning, all methods often face difficulty in precisely editing the image. However, even in such cases as well IABEdit performs better in terms of editing the image while preserving the non-editable regions of the original image. This is validated through the outperforming HM scores in various semantic types of editing instructions, including Direct Simple Instructions (42.00), Morphological Instructions (41.74), Syntactic Instructions (42.12), and achieving competitive results on Lexical Instructions (41.79). While IABEdit significantly and consistently outperforms the very recent SuperEdit method. These results highlight the efficacy of IABEdit in aligning the semantics with the image edits. Wilcoxon signed-rank tests on RealEdit ($n=560$) on simple editing instructions confirm significance ($p<0.05$): \textcolor{black}{IABEdit achieves effect sizes of $r=0.33$ against InstructPix2Pix ($p=1.625\times 10^{-19}$) and $r=0.14$ against MagicBrush ($p=1.238\times 10^{-7}$). In contrast, Reward-InstructPix2Pix shows a smaller effect ($r=0.09, p=0.0103$). These results confirm that our improvements are both statistically significant and practically meaningful.} Figure \ref{fig:complex_editing_qulaitative} shows the qualitative visual results, illustrating how the editing generations vary as the instructions range from simple to complex.


IABEdit understands the instruction semantics and performs precise edits even in many complex instruction scenarios (Figure \ref{fig:complex_editing_qulaitative} (left part)), but in certain cases where words include "devoid", "off" and "less" (Figure \ref{fig:complex_editing_qulaitative} (right part)), both the models fail to understand the instructions.

\section{Additional Quantitative Results}
\label{sec:quant_rebuttal_results}
This section shows additional results on the RealEdit benchmark and EMU Edit test set.

\subsection{Masked CS-P evaluation on RealEdit dataset}
\label{sec:masked_cs_p}
\textcolor{black}{IABEdit and SuperEdit achieve masked CS-P scores of \textit{99.68} and \textit{99.60}, respectively. Masks are generated for instructions with known edit regions, such as "Remove", "Erase", "Replace", "Object Transform", or "Change Region". However, masked-CS-P is less precise for "Add" or "Insert" instructions, since the original image lacks the object, a reference mask cannot be accurately established, given the vast potential for varied model localizations.}

\subsection{Results on EMU Edit test set}
\label{sec:emuedit_test_set_results}

\begin{table}[t!]
\centering
\caption{\label{tab:emuedit_testset} Results on EMU Edit test set.}
\resizebox{0.8\linewidth}{!}{
\begin{tabular}{lcccc}
\hline
\textbf{Method} & \textbf{CLIP}$_{dir}$ $\uparrow$ & \textbf{CLIP}$_{img}$ $\uparrow$ & \textbf{DINO $\uparrow$} & \textbf{CLIP}$_{out}$ $\uparrow$ \\
\hline
UltraEdit & 8.39 & 81.29 & 74.53 & 23.11 \\
EMU Edit & 10.37 & 82.89 & 81.04 & 23.65 \\
MagicBrush & 9.00 & 83.80 & 77.60 & 22.20 \\
\hline
\rowcolor{lightred} IABEdit (+Diffusion) & 9.01 & 85.66 & 83.78 & 24.03 \\
\rowcolor{lightred} IABEdit (+FLUX.1) & \textbf{11.21} & \textbf{88.81} & \textbf{91.02} & \textbf{25.06} \\
\hline
\end{tabular}}
\end{table}

\textcolor{black}{Table \ref{tab:emuedit_testset} presents results on the challenging EMU Edit test set benchmark. CLIP$_{dir}$ measures whether the semantic change in the edited image follows the intended editing instruction, while CLIP$_{out}$ evaluates the alignment between the edited image and the target caption. CLIP$_{img}$ measures similarity between the edited and input images, whereas DINO evaluates structural content preservation between the input source and generated edited images.}

\textcolor{black}{
IABEdit (+FLUX.1) achieves the best performance across all metrics. It attains the highest CLIP$_{dir}$ score (11.21), indicating more accurate instruction-following edits, and the highest CLIP$_{out}$ score (25.06), demonstrating stronger alignment with the target description. It also achieves the highest CLIP$_{img}$ (88.81) and DINO (91.02) scores, showing that the edited images preserve the original content and structural details while performing the desired modifications. Compared with the strongest baseline (EMU Edit), IABEdit (+FLUX.1) improves CLIP$_{dir}$ by +0.84, CLIP$_{out}$ by +1.41, CLIP$_{img}$ by +5.92, and DINO by +9.98, demonstrating that gradient-aligned semantic supervision effectively balances precise instruction following with faithful content preservation.}

\section{Discussion on Image Quality}
\label{sec:image_quality_comparison}
\textcolor{black}{IABEdit does not improve the image quality (see Table 2 in the main paper), as the image editing task is to maintain the quality of the edited image the same as the original, without any lighting or appealing improvements. Thus, our overall evaluation focused on semantic alignment measured through instruction following, and preserving the context (including quality), proven through the preservation score of 4.19 (best) (same metric for baseline).}

\section{Training Cost}
\label{sec:training_cost}

\textcolor{black}{
Training IABEdit method leads to 9.03\% increase in FLOPs (SD+LLAVA) ($\sim$ 6624 GFLOPs) as compared to SuperEdit ($\sim$ 733 GFLOPs). While editing methods like SmartEdit incur higher GFLOPs than IABEdit, it includes two-stage training with additional components, such as Q-Former, Bidirectional Interaction Module and LLaVA-13B with LoRA. There are only 95.68 M extra trainable parameters introduced in IABEdit as compared to SuperEdit (59.53 M). IABEdit performs VLM-free testing, taking 3s/image on a single GPU for 30 timesteps, which is similar to SuperEdit.}

\section{Limitations and Failure Modes}
\label{sec:limitations}

\textcolor{black}{
The failure cases are shown in Figure \ref{fig:complex_editing_qulaitative} and also discussed in Section $5.3$ for complex editing instruction scenarios. Failure cases occur when editing instructions include words like "devoid", "off", and "less", the model is unable to understand them and makes incorrect edits. Another limitation of our method is the increased computational overhead during training due to the use of VLM-based gradient supervision.}


{
    \small
}

\end{document}